\documentclass[10pt,twocolumn,twoside]{IEEEtran}

\usepackage{cite}

\ifCLASSINFOpdf
   \usepackage[pdftex]{graphicx}
\else
\fi
\usepackage[cmex10]{amsmath}
\usepackage{amssymb,amsthm}
\usepackage{algorithmic}
\usepackage{algorithm,bm}
\usepackage{enumerate}
\usepackage{stmaryrd}

\usepackage{color}
\usepackage{mdwmath}
\usepackage{mdwtab}
\usepackage[caption=false,font=footnotesize]{subfig}

\theoremstyle{plain}

\newtheorem{theorem}{Theorem}
\newtheorem{lemma}{Lemma}

\newtheorem{corollary}{Corollary}
\theoremstyle{definition}
\newtheorem{definition}{Definition}
\theoremstyle{remark}
\newtheorem{remark}{Remark}
\newtheorem{example}{Example}

\newcommand{\Matrix}[1]{\mathbf{#1}}

\newcommand{\R}[1]{\mathbb{R}^{#1}}

\DeclareMathOperator{\atan}{arc\,tan}
\DeclareMathOperator{\Orth}{\Cfrak_\mathrm{orth}}
\DeclareMathOperator{\SO}{\Cfrak_\mathrm{orth}}
\DeclareMathOperator{\ST}{\Cfrak_\mathrm{St}}

\newcommand{\GOLrev}[1]{#1}
\newcommand{\RGrev}[1]{#1}
\newcommand{\FBrev}[1]{#1}

\def\sdim{m} 
\def\cdim{d}  
\def\nsamp{n} 
\def\cvdim{h} 
\def\coeff{\boldsymbol{\alpha}}
\def\Coeff{\mathbf{A}}

\def\signal{\mathbf{x}}
\def\Signal{\Matrix{X}}
\def\db{\mathbf{d}}
\def\Db{\Matrix{D}}
\def\Lcal{\mathcal{L}}

\def\Ncal{\mathcal{N}}
\def\lzero{\ell^{0}}
\def\Ub{\Matrix{U}}
\def\deltab{\boldsymbol{\delta}}
\def\Deltab{\boldsymbol{\Delta}}
\def\Afrak{\mathfrak{A}}
\def\Cfrak{\mathfrak{D}}
\def\Pfrak{\mathfrak{P}}
\def\Ck{\Cfrak_{\mathrm{LRIP}}(k,\delta)}
\def\Sphere{\mathfrak{S}}

\def\PP{\mathbb{P}}
\def\EE{\mathbb{E}}
\def\defin{\triangleq}

\def\Id{\mathbf{Id}}

\def\noise {\boldsymbol{\epsilon}}
\def\zb{\mathbf{z}}
\def\Ab{\mathbf{A}}
\def\fro{F}

\font\dsrom=dsrom10 scaled 1200
\newcommand{\indicator}[1]{\textrm{\dsrom{1}}_{#1}}
\def\Kcal{\mathcal{K}}
\def\radius{R}
\def\chisparse {\chi_{k\textrm{-sparse}}}
\def\chiNMF {\chi_{\textrm{NMF}}}
\def\CNMF {\Cfrak_{\textrm{NMF}}}

\def\Lipdev {\Lambda_{\nsamp}} 
\def\Expdev {\Gamma_{\nsamp}} 
\newcommand{\SET}[1]{\llbracket 1; #1\rrbracket}

\makeatletter
\newlength \figwidth
\if@twocolumn
\else
\fi
\makeatother
\begin{document}

\title{Sample Complexity  of Dictionary Learning \protect \\ and other Matrix Factorizations}

%
%
%

\author{
R{\'e}mi Gribonval,~\IEEEmembership{IEEE Fellow},
Rodolphe Jenatton,
Francis Bach,
        Martin~Kleinsteuber, 
				Matthias~Seibert
   \thanks{R. Gribonval heads the PANAMA project-team (Inria \& CNRS), Rennes, France. This work was partially supported by the EU FP7, SMALL project, FET-Open grant number 225913, and by the European Research Council, PLEASE project (ERC-StG-2011-277906).\hfill  remi.gribonval@inria.fr 
   }%
     \thanks{R. Jenatton contributed during his Ph.D. with the Inria - SIERRA Project-Team, Laboratoire d'Informatique de l'Ecole Normale Sup{\'e}rieure, Paris, France and his postdoc at CMAP, Ecole Polytechnique, UMR CNRS 7641, Palaiseau, France. He is now with Amazon, 
Berlin, Germany. \hfill jenatton@amazon.com}%
        \thanks{F. Bach heads the Inria - SIERRA Project-Team, Laboratoire d'Informatique de l'Ecole Normale Sup{\'e}rieure, Paris, France. \hfill francis.bach@ens.fr}%
\thanks{M. Kleinsteuber \& M. Seibert are with the Department of Electrical and Computer Engineering, Technische Universit\"at M\"unchen,  Munich, Germany. This work has been supported by the German DFG funded Cluster of Excellence CoTeSys - Cognition for Technical Systems, and by the DFG grant KL 2189/8-1.
\hfill \{kleinsteuber, m.seibert\}@tum.de 
}
\thanks{This paper was presented in part at SSP 2014 - the IEEE Workshop on Statistical Signal Processing, Jupiters, Gold Coast, Australia.}
\thanks{Copyright (c) 2014 IEEE. Personal use of this material is permitted.  However, permission to use this material for any other purposes must be obtained from the IEEE by sending a request to pubs-permissions@ieee.org}
}

%
%

\markboth{IEEE Transactions on Information Theory}%
{Gribonval \MakeLowercase{\textit{et al.}}: Sample Complexity of Dictionary Learning and other Matrix Factorizations}
%

\maketitle

\begin{abstract}
Many modern tools in machine learning and signal processing, such as sparse dictionary learning, principal component analysis (PCA), non-negative matrix factorization (NMF), $K$-means clustering, etc., rely on the factorization of a matrix obtained by concatenating high-dimensional vectors from a training collection. While the idealized task would be to optimize the expected quality of the factors over the underlying distribution of training vectors, it is achieved in practice by minimizing an  empirical average over the considered collection. The focus of this paper is to provide sample complexity estimates to uniformly control how much the empirical average deviates from the expected cost function. Standard arguments imply that the performance of the empirical predictor also exhibit such guarantees. The level of genericity of the approach encompasses several possible constraints on the factors (tensor product structure, shift-invariance, sparsity \ldots), thus providing a unified perspective on the sample complexity of several widely used matrix factorization schemes.
\GOLrev{The derived generalization bounds behave proportional to $\sqrt{\log(n)/n}$ w.r.t.\ the number of samples $n$ for the considered matrix factorization techniques.}

\end{abstract}

\begin{IEEEkeywords}
Dictionary learning, sparse coding, principal component analysis, $K$-means clustering, non-negative matrix factorization, structured learning, sample complexity.
\end{IEEEkeywords}

%
\IEEEpeerreviewmaketitle

\section{Introduction}\label{sec:intro}

\IEEEPARstart{T}{he} fact that a signal $\signal \in \R{\sdim}$ 
which belongs to a certain class 
has a representation over some class dependent dictionary $\Db \in \R{\sdim \times \cdim}$ is the backbone of many successful signal reconstruction and data analysis algorithms \cite{Mallat:2008aa,sp:elad:2010,Elad:2010wo}. 

That is, $\signal$ is the linear combination of columns of $\Db$, referred to as \emph{atoms}. Formally,
this reads as
\begin{equation}\label{eq:dict_approx}
\signal = \Db \coeff,
\end{equation}
where the coefficient vector $\coeff \in \R{\cdim}$ as well as the dictionary $\Db$ are subject to some constraints. Such a setting covers prominent examples like Principal Component Analysis (PCA), where $\Db$ has orthogonal columns, thus representing the subspace where the signal in the given class is contained.
Another example is the sparse synthesis model, also known as sparse coding, where typically $\Db$ consists of normalized columns that form an overcomplete basis of the signal space, and $\coeff \in \R{\cdim}$ is assumed to be sparse. 

The task of learning such dictionaries from a given set of training data is related to matrix factorization problems. Important examples include Higher-Order SVD (also known as multilinear SVD) \cite{tucker:1966}, sparse coding also called dictionary learning \cite{Olshausen:1997wz,dl:engan:1999,dl:delgado:2003,dl:aharon:2006,dl:mairal:2010,dl:tosic:2011,Rubinstein:2010aa}, its variants with separable \cite{cvpr:hawe:2013} or sparse \cite{dl:rubinstein:2010} dictionaries, Non-negative Matrix Factorization (NMF) \cite{seung:2001algorithms}, $K$-means clustering \cite{Gersho:1992wy}, sparse PCA \cite{dAspremont:2004vv,witten:2009penalized,Zhang:2012uu}, and more. The learning task is expressed formally as follows.

Let $\Signal = [\signal_1,\ldots,\signal_\nsamp] \in \R{\sdim \times \nsamp}$ be the matrix containing the $\nsamp$ training samples arranged as its columns, and let $\Coeff = [\coeff_1,\ldots,\coeff_\nsamp] \in \R{\cdim \times \nsamp}$ contain the corresponding $\nsamp$ coefficient vectors, a common approach to the dictionary learning process is the optimization problem
\begin{equation}\begin{split}\label{eq:basic_dict}
&\operatorname*{minimize}\limits_{\Coeff,\Db}\ \tfrac{1}{2} \|\Signal-\Db \Coeff\|_F^2 + \sum_{i=1}^{\nsamp} g(\coeff_{i})
\ \operatorname{s.t.~}  \ \Db \in \Cfrak.
\end{split}\end{equation}
Therein, $g \colon  \R{\cdim} \to \R{}$ is a function that promotes the constraints for the coefficient vectors, e.g., sparsity or positivity, and $\Cfrak$ is some predefined admissible set of solutions for the dictionary. Note that for PCA there are no constraints on the coefficient vectors, 
which implies the penalty function $g\equiv 0$ in this case.

A fundamental question in such a learning process is the sample complexity issue.  Assuming that the training samples are drawn according to some distribution $\PP$ representing the class of signals of interest, one would ideally like to select the dictionary $\Db^{\star}$ yielding the minimum expected value of~\eqref{eq:basic_dict}. However, having only access to $n$ training samples, one can at best select an empirical minimizer $\hat{\Db}$. Is this empirical minimizer useful beyond the training set from which it has been selected? This depends on how much the empirical cost function deviates from its expectation. 

State of the art sample complexity estimates \cite{MaurerPontil,vainsencher:2010} primarily consider the case where $g$ is the indicator function of a set, such as an $\ell^{1}$ or an $\ell^{0}$ ball, 
$\Cfrak$ is the set of all unit norm dictionaries or a subset with a restricted isometry property, and the distribution $\PP$ is in the class $\Pfrak$ of distributions on the unit sphere of $\R{\sdim}$. 
We generalize these results to:
\begin{itemize}
	\item {\bf General classes of penalty functions.} Examples 
	covered by our results include: the $\ell^{1}$ penalty and its powers; any mixed norm \cite{jenatton:inria-00377732} or quasi-norm \cite{chartrand:2007exact}; the characteristic function of the $\ell^{1}$-ball, of the set of $k$-sparse vectors \cite{vainsencher:2010}, or of non-negative vectors \cite{seung:2001algorithms}. 
	\item {\bf Various classes of dictionaries $\Cfrak$ that can incorporate structures.} Examples covered include: dictionaries with unit norm atoms which are used in many dictionary learning schemes, e.g., K-SVD \cite{dl:aharon:2006}; sparse dictionaries \cite{dl:rubinstein:2010}; shift-invariant dictionaries \cite{mailhe:2008shift}; tensor product dictionaries \cite{cvpr:hawe:2013}; orthogonal dictionaries; non-negative dictionaries \cite{seung:2001algorithms}; topic models \cite{Jenatton:2011vv}; and tensor products of Stiefel matrices used for Higher-order SVDs \cite{tucker:1966,de2000multilinear}. 
	\item {\bf Various classes $\Pfrak$ of probability distributions $\PP$.} Examples include distributions on the unit sphere which are tackled in \cite{vainsencher:2010}, but also certain distributions built using sub-Gaussian random variables \cite{jenatton:hal-00737152}. For more information on sub-Gaussian random variables, see \cite{Buldygin2000}.
\end{itemize}

After formalizing the problem and setting our notations in Section~\ref{sec:pbstatement}, we state our main results in Section~\ref{sec:Main} and illustrate them with families of examples in Section~\ref{sec:Examples}. 
The obtained sample complexity estimates depend on the Lipschitz constant of the optimized cost function~\eqref{eq:basic_dict}, seen as a function of the unknown dictionary $\Db$. This Lipschitz constant is essentially driven by the penalty function $g$, as discussed in Section~\ref{sec:Lipschitz}. Our results rely on the assumption that the distributions $\PP$ in the class $\Pfrak$ satisfy a concentration of measure property, and the structured constraint set $\Cfrak$ plays a role in the sample complexity estimates through its covering number, as discussed in Section~\ref{sec:Concentration} where the main theoretical results are established. Covering numbers for a number of structure sets $\Cfrak$ are considered in Section~\ref{sec:Covering}. Section~\ref{sec:WorkedExamples} gathers worked examples in relation to previous work. Finally, Section~\ref{sec:sample_alt} discusses sharper sample complexity estimates in the high-dimensional setting $\sdim \gg \cdim$.

\section{Notations}
\label{sec:pbstatement}
Sets are denoted with gothic face as $\Cfrak,\Pfrak,\Afrak$. Matrices are written as boldface capital letters like $\mathbf{X},\mathbf{D}$, column vectors are denoted by boldfaced small letters, e.g., $\coeff, \mathbf{d}$ whereas scalars are either capital or small letters like $n,N$. By $v_i$ we denote the $i^{\mathrm{th}}$ element of the vector $\mathbf{v}$, $v_{ij}$ denotes the $i^{\mathrm{th}}$ element in the $j^{\mathrm{th}}$ column of a matrix $\mathbf{V}$, while the $i^{\mathrm{th}}$ column is referred to by $\mathbf{v}_i$. 
The Frobenius norm for matrices is denoted $\|\cdot\|_{F}$ and the corresponding inner product between matrices is denoted $\langle \cdot,\cdot\rangle_{F}$.
The operator norm $\|\cdot\|_{1 \to 2}$ of an $\sdim\! \times\! \cdim$ matrix $\Deltab = [\deltab_{1},\ldots,\deltab_{\cdim}]$ is 
\begin{equation*}
\|\Deltab\|_{1 \to 2} \defin \sup_{\|\coeff\|_{1} \leq 1} \|\Deltab \coeff\|_{2} = \max_{1 \leq i \leq \cdim} \|\deltab_{i}\|_{2}.
\end{equation*}
Finally, $\log$ is the natural logarithm so that $e^{\log t} = t$ for $t>0$.

Given a dictionary $\Db \in \R{\sdim \times \cdim}$ that fulfills certain structural properties and a signal $\signal \in \R{\sdim}$, a representation vector $\coeff \in \R{\cdim}$ is typically obtained by solving the minimization problem
\begin{align*}
	\hat{\coeff} \in \arg \min_{\coeff \in \R{\cdim}} \tfrac{1}{2}\|\signal-\Db\coeff \|_2^2 + g(\coeff),
\end{align*}
where $g: \R{\cdim} \to \R{+} \cup \{+\infty\}$ is a penalty function promoting constraints for the coefficient vector.
For our purposes, the question of whether a minimizer $\hat{\coeff}$ actually exists is irrelevant, and we define the quality of how well a signal $\signal$ can be  coded by a dictionary $\Db$ by
\begin{align*}
	f_\signal(\Db)&\defin \inf_{\coeff \in \R{\cdim}} \Lcal_{\signal}(\Db,\coeff),
\intertext{with}
%
%
	\Lcal_{\signal}(\Db,\coeff)&\defin\tfrac{1}{2}\|\signal-\Db\coeff \|_2^2 + g(\coeff).
\end{align*}
Given $\nsamp$ training samples $\Signal = [\signal_1,\ldots,\signal_\nsamp] \in \R{\sdim \times \nsamp}$, the average quality of how well a dictionary allows a representation for $\Signal$ while considering the constraints imposed on the coefficients is
\begin{align*}
F_\Signal(\Db)\defin \tfrac{1}{\nsamp}\sum_{i=1}^{\nsamp} f_{\signal_i}(\Db).
\end{align*}
The cost function $F_{\Signal}$ can be written in short form as $F_{\Signal}(\Db) = \inf_{\Coeff \in \R{\cdim \times \nsamp}} \Lcal_{\Signal}(\Db,\Coeff)$ with
\begin{equation*}
\Lcal_{\Signal}(\Db,\Coeff) \defin  \tfrac{1}{2\nsamp} \|\Signal-\Db \Coeff\|_{F}^{2}+G(\Coeff)
\end{equation*}
where $\Coeff \defin [\coeff_{1},\ldots,\coeff_{\nsamp}]$ and $G(\Coeff) \defin \frac{1}{\nsamp} \sum_{i=1}^{\nsamp}g(\coeff_{i})$.


\section{Main results \& Outline of our Approach}
\label{sec:Main}

\RGrev{The main contribution of this paper is a general framework to establish the uniform convergence of the function $F_{\Signal}(\cdot)$ to its expectation when the samples $\signal_i$ are all drawn independently according to an unknown probability distribution $\PP$.
We show in this work that 
\begin{equation*}
	\sup_{\Db \in \Cfrak} \left| F_{\Signal}(\Db)-\EE_{\signal \sim \PP} f_{\signal}(\Db) \right| \leq \eta_{\nsamp}(g,\Cfrak,\Pfrak).
\end{equation*}
holds with 
``overwhelming'' probability (that will be controlled explicitly in due time as well as the precision bound $\eta_{\nsamp}(g,\Cfrak,\Pfrak) \propto \sqrt{\log \nsamp/\nsamp}$)}.

\RGrev{
A particular consequence is in terms of generalization bound. An ideal objective in dictionary learning and related matrix factorization problems would be to select the optimal dictionary for the underlying distribution $\PP$ of training samples, 
\begin{equation*}
\Db^{\star} \in \arg\min_{\Db \in \Cfrak} \EE_{\signal\sim \PP}f_\signal(\Db).
\end{equation*}
In practice, one can at best access an empirically optimal dictionary $\hat{\Db}_{\nsamp} \in \Cfrak$, which minimizes $F_{\Signal}(\Db)$ given $\nsamp$ training samples gathered in $\Signal$. How close is its performance to that of $\Db^{\star}$? A consequence of our main result is a generalization bound for the empirical optimum $\hat{\Db}_{n}$: with controlled high probability, 
\begin{equation*}
 \EE_{\signal\sim \PP}f_\signal(\hat{\Db}_{\nsamp}) \leq  \EE_{\signal\sim \PP}f_\signal(\Db^{\star}) + 2\eta_{\nsamp}.
\end{equation*}
}

\RGrev{Note that the uniform convergence result holds for all dictionaries $\Db \in \Cfrak$, and not only at the global optimum of the learning problem. 
Moreover, even though finding the empirical minimizer $\hat{\Db}_{\nsamp}$ often means facing a difficult non-convex optimization problem, recent work on sparse dictionary learning establishes that certain polynomial time algorithms can provably find it with high probability in certain scenarii \cite{Spielman:2012ue,Arora:2013vq,Agarwal:2013tya}. Uniform convergence results as established here should enable further investigation of the finite sample analysis of such algorithms.}

\RGrev{The main result, formally stated in Theorem~\ref{th:MainTheorem} in Section~\ref{sec:maintheorem}, combines a standard argument based on covering numbers with an estimation of the expected Lipschitz constant of the function $\Db \mapsto F_{\Signal}(\Db)$. Our main technical contribution is the estimation of this Lipschitz constant, which is summarized in Theorems~\ref{th:LipschitzNormPenalties} and~\ref{th:LipschitzKSparse} in Section~\ref{sec:mainLipschitz}.}
\GOLrev{To prove the main result, we first show that $\Db \mapsto F_{\Signal}(\Db)$ is Lipschitz with a constant expressed in terms of $\Signal$ and $g$. To be more precise, we prove the property
$|F_\Signal(\Db) - F_\Signal(\Db^\prime)| \leq L \cdot \|\Db - \Db^\prime\|_{1 \to 2}$
for all admissible dictionaries $\Db,\Db^\prime \in \Cfrak$, where $L$ is dependent on $\Signal$ and $g$. In the course of our work we provide more tangible bounds for the Lipschitz constant. However, this requires certain assumptions on the penalty function $g$. These assumptions will be motivated and introduced in the remainder of this section. Therefore, we refrain from presenting the concrete bounds at this point. They can be found in Corollary~\ref{le:ConstantsWithNormPenalties} and Lemma~\ref{le:ConstantsWithKSparse}.
The discussion is concluded using an argument based on covering numbers and concentration of measure.
While the latter technique is fairly standard, a 
major contribution of this paper lies in the identification of two large classes of penalty functions for which the desired Lipschitz property holds and is nicely controlled.
In the following, we will give a detailed insight into the required components to provide the generalization bound. Furthermore, we motivate the assumptions that we enforce on the penalty function $g$ (these are labeled with \textbf{A}), or jointly on the penalty function $g$ and the set of admissible dictionaries $\Cfrak$ (labeled with \textbf{B}), and on the probability distribution (labeled with \textbf{C}).}

\subsection{Role of the constraint set $\Cfrak$}
The structure of the constraint set $\Cfrak$ is incorporated in the analysis by employing a standard $\epsilon$-net argument. 
A compact constraint set $\Cfrak \subset \R{\sdim \times \cdim}$ can be covered with a finite number of balls of small radius, i.e., it has a finite covering number 
\begin{equation*}
	\mathcal{N}(\Cfrak,\epsilon) 
	\defin \min\{ \sharp \mathfrak{Q} \,:\, \mathfrak{Q} \subset \R{\sdim \times \cdim}, \Cfrak \subset \bigcup_{\mathbf{q} \in \mathfrak{Q}} B_\epsilon(\mathbf{q}) \}
\end{equation*}
where $\sharp$ denotes the cardinality of a set. For further background about covering numbers, we refer the reader to~\cite{Cucker2002} and references therein. 
In our setting the $\epsilon$-balls $B_\epsilon(\cdot)$ are defined with respect to the metric $\|\cdot\|_{1\to2}$ since the unit ball in this metric is closely connected with the common constraint that a dictionary has unit norm columns.

We postpone explicit estimates of covering numbers to Section~\ref{sec:Covering}, but notice that covering numbers for all the considered classes satisfy the generic upper bound
\begin{equation}
\label{eq:ExplicitGenericCoveringAssumption}
\Ncal(\Cfrak,\epsilon) \leq \left(C/\epsilon\right)^{\cvdim},\ \forall\ 0<\epsilon\leq 1
\end{equation}
where $\cvdim\geq 1$ is a measure of the ``dimension'' of $\Cfrak$, and $C\geq 1$. In particular, for the set of all dictionaries in $\R{\sdim \times \cdim}$ with unit-norm columns, which is one of the most common constraint sets, these constants take the values $\cvdim = \sdim \cdim$ and $C=3$. The sample complexity will be essentially driven by the ``dimension'' $\cvdim$, while the constant $C$ will only play a logarithmic role.

\subsection{Lipschitz property for ``norm-like'' penalty functions $g$}

In short, we will prove in Section~\ref{sec:Lipschitz} that under assumptions 
\begin{itemize}
\item{\bf A1:} $g$ is non-negative;
\item{\bf A2:} $g$ is lower semi-continuous;
\item{\bf A3:} $g$ is coercive: $g(\coeff) \to \infty$ as $\|\coeff\| \to \infty$;
\end{itemize}
the function $\Db \mapsto F_{\Signal}(\Db)$ is Lipschitz over $\R{\sdim \times \cdim}$ with controlled constant. With the additional assumption
\begin{itemize}
\item {\bf A4:} $g(\mathbf{0}) = 0$;
\end{itemize}
the Lipschitz constant is bounded by
\begin{equation}
\label{eq:DefLipschitzConstantUniform}
L_{\Signal}(\bar{g}) \defin \tfrac{1}{\nsamp} \sum_{i=1}^{\nsamp} \|\signal_{i}\|_{2} \cdot \bar{g}(\|\signal_{i}\|_{2}^{2}/2)
\end{equation}
where we define the auxiliary function $\bar{g}$ as follows.
\begin{definition}\label{def:DefCoercivityFn}
For penalty functions $g$ satisfying {\bf A1-A3}, 
\begin{equation*}
\bar{g}(t) \defin \sup_{\substack{\coeff \in \R{\cdim} \\ g(\coeff)\leq t}} \|\coeff\|_{1},\ t \geq 0.
\end{equation*}
\end{definition}
\noindent The presence of the $\ell^{1}$ norm in this definition results from the choice of the metric $\|\cdot\|_{1 \to 2}$ to measure covering numbers. It will become clearer in the proof of Lemma~\ref{le:ConstantsWithNormPenalties_A1A3} where we need to control $\|\coeff\|_{1}$ where $\coeff$ is a (near) minimizer of $\Lcal_{\signal}(\Db,\cdot)$.

The following properties of this function will be useful.
\begin{lemma}\label{le:Enveloppe}
The function $\bar{g}$ as defined in Definition~\ref{def:DefCoercivityFn} is non-increasing and takes its values in $\R{+}$.
\end{lemma}
\begin{IEEEproof}[Proof of Lemma~\ref{le:Enveloppe}]
The sublevel sets of $g$ are nested, hence $\bar{g}$ is non-decreasing. 
Since $g$ is coercive, its sublevel sets are bounded, hence $\bar{g}$ takes its values in $\R{+}$.
\end{IEEEproof}

Note that Assumption~\textbf{A4} is a convenience that should not be taken too literally, as penalty functions that fulfill \textbf{A1}-\textbf{A3} and have a global minimum at $\mathbf{0}$ can be manipulated to fulfill \textbf{A4} by subtracting the value of $g$ at $\mathbf{0}$.

\subsection{Lipschitz property under joint assumptions on $g$ and $\Cfrak$}
While Assumptions {\bf A1-A4} cover a wide range of penalty functions, they do not cover popular penalties related to the $\lzero$ quasi-norm, such as the indicator function (also called characteristic function) of $k$-sparse vectors:
\begin{equation*}
\chisparse(\coeff)
\defin
		\begin{cases}
		0,& \text{if } \|\coeff\|_{0} \leq k,\\
		+\infty, & \text{otherwise};
	\end{cases}
\end{equation*}
Vainsencher {\em et al.} \cite{vainsencher:2010} deal with the latter case under an incoherence assumption on $\Db$, i.e., by restricting $\Cfrak$ to be the class of dictionaries with small (cumulative) coherence. A careful study of their technique shows that the results are actually valid under an assumption related to the well-known restricted isometry property (RIP) \cite{CS:candes:2006a,Elad:2010wo}.
In fact, while the RIP is usually expressed in its symmetric form, only its lower bound actually plays a role in the considered context.  This justifies the following definition.
\begin{definition}[Lower RIP]
\label{def:LowerRIP}
For $k \leq \min(\sdim,\cdim)$, we denote $\delta_{k}(\Db)$ the smallest $0 < \delta \leq 1$ such that for all $k$-sparse vectors $\coeff$
\begin{align*}
(1-\delta)\|\coeff\|_2^2 \leq \|\Db \coeff \|_2^2.
\end{align*}
For $k \leq \min(\sdim,\cdim)$ and $0<\delta<1$ we define the compact set
\begin{equation}
\label{eq:DefRIPClass}
\Ck \defin \left\{ \Db \in \R{\sdim \times \cdim} \,:\, \delta_{k}(\Db) \leq \delta\right\}.
\end{equation}
\end{definition}
Similarly, Assumptions {\bf A1-A4} do not cover the indicator function of non-negative coefficients, 
\begin{equation*}
\chiNMF(\coeff)
\defin
		\begin{cases}
		0,& \text{if } \alpha_{i} \geq 0, \forall 1 \leq i \leq \cdim,\\
		+\infty, & \text{otherwise};
	\end{cases}
\end{equation*}
which is typically used in conjunction with a non-negativity constraint on $\Db$ as well, namely $\Db \in \CNMF$, where
\begin{align}
\CNMF \defin \Big\{& \Db \in \R{\sdim \times \cdim} \,:\,
	 d_{ij} \geq 0,\ \forall i,j\Big\}.\label{eq:DefCNMF}
\end{align}

To unify the treatment of such penalty functions not covered by Assumptions {\bf A1-A4}, we develop complementary results based on joint assumptions on $g$ and the constraint set $\Cfrak$: 
\begin{itemize} 
\item {\bf B1:} $g = \chi_{\mathcal{K}}$ is the indicator function of a set $\mathcal{K}$;
\item {\bf B2:} there is $\kappa >0$ such that for any $\coeff \in \mathcal{K}$ and $\Db \in \Cfrak$, 
\begin{equation*}
	\kappa \|\coeff\|_{1}^{2} \leq \|\Db\coeff\|_{2}^{2}
\end{equation*}
\item {\bf B3:} $\mathcal{K}$ contains the origin: $\mathbf{0} \in \mathcal{K}$.
\end{itemize}
Note that {\bf B2} is a type of restricted eigenvalue condition, see e.g., \cite[Equation (30)]{Negahban:2012en}.

These assumptions hold for $g = \chisparse$ and $\Cfrak = \Ck$ with $\kappa = (1-\delta)/\cdim$, and they will be shown to also hold for $g = \chiNMF$ and $\Cfrak = \CNMF(\kappa) \subset \CNMF$ where
\begin{align}
\CNMF(\kappa) \defin \Big\{& \Db \in \R{\sdim \times \cdim} \,:\,
	 d_{ij} \geq 0,\ \|\db_{j}\|_{2}^{2} \geq \kappa \cdim,\ \forall i,j\Big\}.\label{eq:DefCNMFkappa}
\end{align}

Under Assumptions {\bf B1-B3} we show that the function $\Db \mapsto F_{\Signal}(\Db)$ is Lipschitz over $\Cfrak$. Its Lipschitz constant can be expressed as $L_{\Signal}(\bar{g})$ (see Eq.~\eqref{eq:DefLipschitzConstantUniform}) where $\bar{g}$ is defined in the context of assumptions {\bf B1-B3} as: 
\begin{definition}\label{def:DefBarGCaseB}
For penalty functions $g$ and $\Cfrak$ satisfying {\bf B1-B3}, \RGrev{we define}
\begin{equation*}
\bar{g}(t) \defin 2 \sqrt{2t/\kappa}
\end{equation*}
\end{definition}
Occasionally, an additional assumption will be helpful:
\begin{itemize}
\item {\bf B4:} $\Cfrak$ is convex.
\end{itemize}

\subsection{Role of the class $\Pfrak$ of probability distributions}

Finally, the results rely on two assumptions on the probability distribution $\PP$ from which the training samples $\signal_{i}$ are assumed drawn i.i.d.:
\begin{itemize}
	\item first, we need to control the Lipschitz constant $L_{\Signal}(\bar{g})$ when the sample size $\nsamp$ is large. 
		\item second, given $\Db$, we need to control the concentration of the empirical average $F_{\Signal}(\Db)$ around its expectation;
\end{itemize}
 By the law of large numbers, the first condition holds under assumption
 \begin{itemize}
 \item {\bf C1:} bounded moment
	\begin{equation*}
	L_{\PP}(\bar{g}) \defin \EE_{\signal \sim \PP}  \|\signal_{i}\|_{2} \bar{g}\left(\tfrac{\|\signal_{i}\|_{2}^{2}}{2}\right) < + \infty.
	\end{equation*}
\end{itemize}
We will see on many examples (Table~\ref{tab:EnveloppeBarG}) that $t\bar{g}(t^{2}/2) \propto t^{2}$ or $t\bar{g}(t^{2}/2) \propto t^{3}$, hence this is a relatively mild condition. From a quantitative perspective our results will be expressed using
\begin{align}
\label{eq:DefProbLipschitz}
\Lipdev(L,\bar{g})
& \defin
\PP\left(\tfrac{1}{\nsamp} \sum_{i=1}^{\nsamp} \|\signal_{i}\|_{2} \bar{g}\left(\tfrac{\|\signal_{i}\|_{2}^{2}}{2}\right) > L\right).
\end{align}
By abuse of notation we will simply write $\Lipdev(L)$ and will exploit the fact that $\lim_{\nsamp \to \infty} \Lipdev(L) = 0$ for $L>L_{\PP}(\bar{g})$.

The second condition is measured through
\begin{align}
\label{eq:DefProbConcentration}
\Expdev(\gamma)
& \defin
\sup_{\Db \in \Cfrak} \PP\left(\left|\tfrac{1}{\nsamp} \sum_{i=1}^{\nsamp} f_{\signal_{i}}(\Db) - \EE f_{\signal}(\Db)\right| > \gamma\right).
\end{align}
As discussed in Section~\ref{sec:Concentration}, our main results exploit {\bf C1} and
\begin{itemize}
\item {\bf C2:}  there are $c>0$ and $T \in (0,+\infty]$  such that
\begin{equation*}
\Expdev(c\tau) \leq 2 \exp(-\nsamp \tau^{2}),\ \forall 0 \leq \tau \leq T,\ \forall \nsamp.
\end{equation*}
\end{itemize}

As shown in Section~\ref{sec:Concentration}, this covers the case of probability distributions on the unit sphere in $\R{\sdim}$ (see, e.g., \cite{vainsencher:2010}), 
\begin{equation*}
	\Sphere^{\sdim-1} \defin \{\signal \in \R{\sdim} \,:\, \|\signal\|_{2}=1\},
\end{equation*}
and more generally on Euclidean balls $B_{R}$ of given radius $\radius$,
\begin{definition}[Probability distributions on a ball]\label{def:ProbOnSphere}
The set of probability distribution within a ball of radius $R$ is given by 
\begin{equation*}
\Pfrak_{B_{\radius}} \defin \left\{\PP \,:\, \PP(\|\signal\|_{2}\leq \radius)=1 \right\}.
\end{equation*}
For $\PP \in \Pfrak_{B_{\radius}}$, {\bf C2} holds with $c=\radius^{2}/\sqrt{8}$, $T=+\infty$, and {\bf C1} holds, with $\Lipdev(L)=0$ for $L=R\bar{g}(R^{2}/2)$ (Lemma~\ref{le:ConcentrationSphere}).
\end{definition}
Assumption {\bf C2} also covers the following classes which contain the sub-Gaussian sparse signal model proposed in \cite{jenatton:hal-00737152}. More details will be given in Section~\ref{sec:Examples}.
\begin{definition}\label{def:SubGaussian}
A distribution is in $\Pfrak_{A}$, $A>0$ if
\begin{equation}
\label{eq:ConcentrationSubGaussian2}
\PP(\|\signal\|_{2}^{2} \geq A t) \leq \exp(-t),\ \forall t \geq 1.
\end{equation}
For $\PP \in \Pfrak_{A}$, {\bf C2} holds with $c=12A$, $T=1$ (Lemma~\ref{le:ConcentrationSubGaussian}), and {\bf C1} holds as soon as $\bar{g}$ has at most polynomial growth. 
\end{definition}

\subsection{Main result}
\label{sec:maintheorem}
Our main result is obtained using a standard union bound argument. The details are in Section~\ref{sec:Concentration}.
In short, under assumptions {\bf A1-A4} or {\bf B1-B3}, together with {\bf C1-C2}, we show:
\begin{theorem}\label{th:MainTheorem}
Consider $L > L_{\PP}(\bar{g})$ and define
\begin{align}
	\label{eq:def_beta_org}
	\beta & \defin h \cdot \max\left(\log \tfrac{2LC}{c},1\right),\\
	\eta_{\nsamp}(g,\Cfrak,\Pfrak) \label{eq:MainSampleComplexity}
	& \defin
	3c \sqrt{\tfrac{\beta \log \nsamp}{\nsamp}} + c \sqrt{ \tfrac{\beta+x}{\nsamp}}.
\end{align}
Then, given $0\leq x \leq \nsamp T^{2}-\beta \log \nsamp$ we have: except with probability at most $\Lipdev(L) + 2 e^{-x}$,
\begin{equation}
\label{eq:UniformConvergence2}
	\sup_{\Db \in \Cfrak} \left| F_{\Signal}(\Db)-\EE_{\signal} f_{\signal}(\Db) \right| \leq \eta_{\nsamp}(g,\Cfrak,\Pfrak).
\end{equation}
Note that $\Lipdev(L)$ is primarily characterized by the penalty function $g$ and the class of probability distributions $\Pfrak$ (see~\eqref{eq:DefProbLipschitz}), while the constants $C,\cvdim \geq 1$ depend on the class of dictionaries $\Cfrak$, see~\eqref{eq:ExplicitGenericCoveringAssumption}, and $c>0$, $0<T\leq \infty$ depend on the class of probability distributions $\Pfrak$, see~{\bf C2}.
\end{theorem}
The constant $3c$ in the first term of the right hand side of~\eqref{eq:MainSampleComplexity} is improved to $2c$ for penalties that satisfy {\bf A1-A4} or {\bf B1-B4}, see Lemma~\ref{le:ExplicitSampleComplexity} in Section~\ref{sec:Concentration} where Theorem~\ref{th:MainTheorem} is proved.

\section{Examples}
\label{sec:Examples}

We now discuss concrete settings covered by our results. More detailed worked examples will be given in Section~\ref{sec:WorkedExamples}.
\subsection{Penalty functions satisfying {\bf A1-A4}}

Many classical penalty functions are covered by Assumptions {\bf A1-A4}: norms, quasi-norms, their powers, indicator functions of compact sets containing the origin, and more. 
\begin{table*}[htdp]
\caption{\label{tab:EnveloppeBarG} Penalty functions $g: \R{\cdim} \to \R{}$ and associated $\bar{g}$ (from either Definition~\ref{def:DefCoercivityFn} or Definition~\ref{def:DefBarGCaseB})}
\begin{center}
\begin{tabular}{|c|c|c|c|}
\hline
$g(\coeff)$ & remark &  $\bar{g}(t)$ & $t\bar{g}(t^{2}/2)$\\
\hline
(quasi)norm $g(\coeff)$ & $C_{g} \defin \sup_{\coeff\neq \mathbf{0}} \|\coeff\|_{1}/g(\coeff)$ & $C_{g}t$ & $\tfrac{C_{g}}{2} \cdot t^{3}$\\
\hline
$\|\coeff / \lambda\|_{p}^{r}$, & $0<p \leq \infty$, $0< r < \infty$ & $\cdim^{(1-1/p)_{+}} \cdot \lambda \cdot t^{1/r}$ & $\cdim^{(1-1/p)_{+}} \cdot \lambda \cdot t \cdot (t^2 / 2)^{1/r}$ \\ 	
\hline
$\chi_{\|\coeff\|_{p} \leq \lambda}$, & $0<p \leq \infty$ & $\cdim^{(1-1/p)_{+}} \cdot \lambda$ & $\cdim^{(1-1/p)_{+}} \cdot \lambda \cdot t$\\
\hline
$\chisparse(\coeff)$ & on $\Ck$, see \eqref{eq:DefRIPClass} & $2 \sqrt{k/(1-\delta)} \sqrt{2t}$ & $2 t^{2 }\sqrt{k/(1-\delta)}$\\
\hline
$\chiNMF(\coeff)$ & on $\CNMF(\kappa)$, see \eqref{eq:DefCNMFkappa} & $2 \sqrt{2t/\kappa}$ & $2  t^{2}/ \sqrt{\kappa}$\\
\hline
\end{tabular}
\end{center}
\end{table*}
%

As a side note, if the penalty function $g$ is invariant by sign flips (this is the case, e.g., for standard norms, mixed norms, etc.), one may notice that $\bar{g}(t)$ is related to the Fenchel conjugate \cite{Boyd2004} of the function $h(\coeff) = \chi_{g(\coeff) \leq t}$, which reads as
\begin{equation*}
h^\star(\mathbf{q}) \defin \sup_{\coeff} \left(\langle \mathbf{q},\coeff\rangle- h(\coeff)\right)
\end{equation*}
since $\bar{g}(t) = h^{\star}(\mathbf{1})$ where $\mathbf{1}$ is the vectors with all ones. 

\subsubsection{Norms and quasi-norms}
All standard $\ell^{p}$ norms ($1 \leq p \leq \infty)$, quasi-norms $(0< p < 1)$, and mixed $\ell^{p,q}$ norms used in structured sparse regularization satisfy {\bf A1-A4}, and we have
$\bar{g}(t) = C_{g} t$
with
\begin{align*}
C_{g} &= \sup_{\coeff \neq \mathbf{0}} \frac{\|\coeff\|_{1}}{g(\coeff)} < \infty
\end{align*}

For example, for $g(\coeff) = \|\coeff\|_{p}$ with $\coeff \in \R{\cdim}$, we have 
\begin{align*}
C_{g} &= \cdim^{(1-1/p)_{+}}
\end{align*}
where $(\cdot)_{+} \defin \max(\cdot,0)$. We let the reader check that for mixed $\ell^{1,2}$ norms, $\|\coeff\|_{1,2} = \sum_{i} \|\coeff_{J_{i}}\|_{2}$, where the sets $J_{i}$ partition the index set $\SET{\cdim}$, we have $C_{g} = j^{1/2}$ where $j = \max_{i} \sharp J_{i}$. 

\subsubsection{Indicator functions of compact sets}
Assumptions {\bf A1-A3} also hold for the indicator function of a compact set $\mathcal{K}$,
\begin{equation*}
g(\coeff) = \chi_{\coeff \in \mathcal{K} } \defin
		\begin{cases}
		0,& \mbox{if}\ \coeff \in \mathcal{K},\\
		+\infty, & \mbox{otherwise}.
		\end{cases}
\end{equation*}
For such penalties we have $\bar{g}(t) = \sup_{\coeff \in \mathcal{K}} \|\coeff\|_{1}$ for all $t \geq 0$. 
Assumption {\bf A4} further requires that the compact set contains the origin, $\mathbf{0} \in \mathcal{K}$.

In particular, assumptions {\bf A1-A4} hold for the indicator function of a ball defined by a (quasi-)norm, i.e., with $\mathcal{K} = \{\coeff: \|\coeff\| \leq \lambda\}$. For an $\ell^{p}$ (quasi-)norm, we have $\bar{g}(t) = \cdim^{(1-1/p)_{+}} \lambda$.

\subsubsection{More examples}

Assumptions {\bf A1-A4} hold when $g$ is a power of a (quasi-)norm, $g(\coeff) = \|\coeff / \lambda\|_{p}^{r}$, with $r>0$, leading to $\bar{g}(t) = \cdim^{(1-1/p)_{+}} \lambda t^{1/r}$.  Note that the indicator function of $\mathcal{K} = \{\coeff : \|\coeff\| \leq \lambda\}$ reads as the limit 
when $r \to \infty$.

There are of course measures that do not fit in our framework, such as the arctangent sparsity measure, which mimics the $\lzero$ one \cite{hage:robust2014} and is defined as
\begin{equation*}
	g_{\operatorname{atan}}(\coeff) \defin \sum_{j=1}^{d} \atan^2\left(\alpha_j\right).
\end{equation*}
It is not coercive and does thereby not meet assumption \textbf{A3}.

\subsection{Penalty functions and constraint sets that satisfy {\bf B1-B3}}
Consider $g(\coeff) = \chisparse(\coeff) = \chi_{\mathcal{K}}$ with the set $\mathcal{K} = \{\coeff: \|\coeff\|_{0} \leq k\}$ and $\Cfrak = \Ck$. Assumptions {\bf B1} and {\bf B3} obviously hold since $\mathcal{K}$ is a finite union of subspaces. Moreover, by definition of $\Ck$, for any $\coeff \in \mathcal{K}$ and $\Db \in \Cfrak$ we have
\begin{equation*}
\|\Db\coeff\|_{2}^{2} \geq (1-\delta) \|\coeff\|_{2}^{2} \geq \tfrac{1-\delta}{k} \|\coeff\|_{1}^{2}
\end{equation*}
where the rightmost inequality follows from the fact that $\coeff$ is $k$-sparse. Hence assumption {\bf B2} holds with $\kappa = (1-\delta)/k$.

Consider now $g(\coeff) = \chiNMF(\coeff) = \chi_{\mathcal{K}}$ with $\mathcal{K} = \{\coeff \,:\, \alpha_{i} \geq 0,\ 1 \leq i \leq \cdim\}$, and $\Cfrak = \CNMF(\kappa)$. Assumptions {\bf B1} and {\bf B3} obviously hold since $\mathcal{K}$ is an orthant. Moreover, since $\Db \in \Cfrak$ has non-negative entries\footnote{In fact, as in \cite{MaurerPontil} we observe it is sufficient to have $\langle \db_{i},\db_{j}\rangle \geq 0$, $i \neq j$}, for any $\coeff \in \mathcal{K}$ we have
\begin{align*}
	\|\Db\coeff\|_{2}^{2} 
	&=
	\sum_{i,j=1}^{\cdim} \alpha_{i} \alpha_{j}  \langle \db_{i},\db_{j}\rangle
	\geq \sum_{i=1}^{\cdim} \alpha_{i}^{2} \|\db_{i}\|_{2}^{2}\\
	&\geq  \|\coeff\|_{2}^{2} \cdot \min_{1 \leq j \leq \cdim} \|\db_{j}\|_{2}^{2} 
	\geq \|\coeff\|_{1}^{2} \cdot \min_{1 \leq j \leq \cdim} \|\db_{j}\|_{2}^{2}/\cdim.
\end{align*}
Hence assumption {\bf B2} holds.

Table~\ref{tab:EnveloppeBarG} summarizes, for standard penalty functions satisfying either {\bf A1-A4} or {\bf B1-B3}, the expression of $\bar{g}$ and that of $t \bar{g}(t^{2}/2)$ which appears in the expression of $L_{\Signal}(\bar{g})$ (see~\eqref{eq:DefLipschitzConstantUniform}).

\subsection{Covering numbers}

Table~\ref{tab:Covering} summarizes the covering numbers of the structured classes of dictionaries considered in Section~\ref{sec:Covering}. 
\GOLrev{The provided covering numbers all depend on the signal dimension $m$, which may lead to sub-optimal sample complexity estimates for high-dimensional problems, i.e., when $\sdim \gg \cdim$. This issue is discussed in section Section~\ref{sec:sample_alt}.
}
\begin{table*}[htdp]
\caption{\label{tab:Covering} Classes of structured dictionaries $\Cfrak$ with covering numbers $\Ncal(\Cfrak,\epsilon) \leq (C/\epsilon)^{\cvdim}$, $0<\epsilon\leq1$. See Section~\ref{sec:Covering}} 
\begin{center}
\begin{tabular}{|c|c|c|}
\hline
$\Cfrak$ & exponent $\cvdim$ & constant $C$\\
\hline
Unit norm $\Cfrak(\sdim,\cdim)$ & $\sdim \cdim$ & $3$\\
\hline 
Separable $\Cfrak_{\mathrm{sep}}$ & $\sum_{i} \sdim_{i} \cdim_{i}$ & $3$\\
\hline
Sparse $\Cfrak_{\mathrm{sparse}}(\sdim,\cdim)$ & $s\cdim$ & $3 \tbinom{\sdim}{s}^{1/s}$\\
\hline
Orthogonal $\Cfrak_{\mathrm{orth}}(\sdim)$ & $\sdim(\sdim-1)/2$ & $3\pi e^\pi$\\
\hline
Stiefel $\Cfrak_{\mathrm{St}}(\sdim,\cdim)$ & $\sdim\cdim-\cdim(\cdim+1)/2$ & $3\pi e^\pi$\\
\hline
Stiefel Tensor $\Cfrak_{\otimes\mathrm{St}}$ & $\sum_i\sdim_i\cdim_i-\cdim_i(\cdim_i+1)/2$ & $3\pi e^\pi$\\
\hline
\end{tabular}
\end{center}
\end{table*}

\subsection{Probability distributions}

While previous work \cite{MaurerPontil,vainsencher:2010} only covers distributions on the unit sphere or in the volume of the unit ball, our results cover more complex models, such as the model of sparse signals with sub-Gaussian non-zero coefficients introduced in \cite{jenatton:hal-00737152}. 

\begin{definition}[Sub-Gaussian model of sparse signals]
\label{def:subgaussianmodel}
Given a fixed reference dictionary $\Db_0 \in \Cfrak$, each noisy sparse signal $\signal \in \R{\sdim}$ is built from the following steps:
\begin{enumerate}
\item  {\em Support generation:} Draw uniformly without replacement $k$ atoms out of the $\cdim$ available in $\Db_0$.
This procedure thus defines a support $J\defin\{j\in\SET{\cdim}\,:\, \delta(j)=1\}$ whose size is $|J|=k$, 
and where $\delta(j)$ denotes the indicator function equal to one if the $j$-th atom is selected, zero otherwise.
Note that $\EE[\delta(j)]=\frac{k}{\cdim}$ and for $i\neq j$ we further have $\EE[\delta(j)\delta(i)]=\frac{k(k-1)}{\cdim(\cdim-1)}$.

\item {\em Coefficient generation:} Define a sparse vector $\coeff_0 \in \R{\cdim}$ supported on $J$ whose entries in $J$ are generated i.i.d.\ according to a {\em sub-Gaussian distribution}: for $j$ not in $J$, $[\coeff_0]_j$ is set to zero; otherwise, we assume there exists some $\sigma > 0$ such that for $j \in J$ we have, for all $t \in \R{}$, $\EE\{ \exp(t[\coeff_0]_{j}) \} \leq \exp (\sigma^{2}t^{2}/2)$ . We denote $\sigma_{\alpha}$ the smallest value of $\sigma$ such that this property holds. For background on sub-Gaussianity, see~\cite{Buldygin2000,Vershynin2010}. 

\item {\em Noise:} Eventually generate the signal $\signal=\Db_0 \coeff_0 + \noise$, 
where the entries of the additive noise $\noise\in \R{\sdim}$ are assumed i.i.d.\ sub-Gaussian with parameter $\sigma_{\varepsilon}$.
\end{enumerate}
\begin{remark}
The model in~\cite{jenatton:hal-00737152} is more restricted: it assumes that, for $j \in J$, $|[\coeff_0]_{j}| \geq \underline{\alpha}$ almost surely, where $\underline{\alpha} > 0$.
\end{remark}
\end{definition}
The distribution $\PP$ of this sub-Gaussian sparse signal model belongs to $\Pfrak_{A}$ with $A = k\sigma_{\alpha}^{2}+\sdim \sigma_{\varepsilon}^{2}$, as shown below (the argument can be originally found in~\cite{jenatton:hal-00737152}).

\begin{lemma}[From~\cite{Hsu2011}]\label{lem:onesided_quadratictail}
Let us consider $\zb \in \R{m}$ a random vector of independent sub-Gaussian variables with parameters
upper bounded by $\sigma>0$. Let $\Ab \in \R{m\times p}$ be a fixed matrix. 
For any $t \geq 1$, we have 
\begin{equation*}
\PP\Big(
\|\Ab\zb\|_2^2 > 5\sigma^2 \|\Ab\|_\fro^2 t 
\Big) \leq \exp(-t).
\end{equation*}
\end{lemma}

\begin{corollary}\label{cor:l2norm_signal}
Let $\signal$ be a signal following the model of Definition~\ref{def:subgaussianmodel}. 
For any $t \geq 1$ we have 
\begin{align*}
\PP\left( \|\signal\|_2^2 > 5(k\sigma_{\alpha}^2 +m\sigma_{\varepsilon}^2 ) t \right) &\leq \exp(-t).
\end{align*}
\end{corollary}
\begin{IEEEproof}
The considered norm can be expressed as follows 
\begin{equation*}
\|\signal\|_{2}^{2}
=
\left\|
\left[
\begin{array}{cc}
\sigma_{\alpha} [\Db_0]_J & \sigma_{\varepsilon}\Id\end{array}
\right]  \binom{\frac{1}{\sigma_{\alpha}}[\coeff_0]_J}{\frac{1}{\sigma_{\varepsilon}}\noise} 
\right\|_2^{2}.
\end{equation*}
The result is a direct application of Lemma~\ref{lem:onesided_quadratictail} conditioned to the draw of $J$, and the observation that
\begin{equation*}
\left\|
\left[
\begin{array}{cc}
\sigma_{\alpha} [\Db_0]_J & \sigma_{\varepsilon}\Id\end{array}
\right]  
\right\|_\fro^{2}
= \|[\Db_0]_J\|_{\fro}^{2} \cdot \sigma_{\alpha}^{2}  + m  \sigma_{\varepsilon}^{2} 
= k \sigma_{\alpha}^{2} + m  \sigma_{\varepsilon}^{2}.
\end{equation*}
The bound being independent of $J$, the result is also true without conditioning. 
\end{IEEEproof}

Section~\ref{sec:WorkedExamples} will detail worked examples with results for specific combinations of dictionary class $\Cfrak$, penalty $g$, and probability distribution $\PP$, relating our results to prior work.

\section{Lipschitz continuity of $F_{\Signal}$}
\label{sec:Lipschitz}
Under appropriate assumptions on the penalty function $g$, we prove below that the function $F_{\Signal}(\Db)$ is Lipschitz continuous with a controlled constant $L_{\Signal}(\bar{g})$. We begin by a one-sided Lipschitz property with an additional quadratic term that we will soon get rid of. 
\begin{lemma}\label{le:OneSidedLipschitz}
Let $\|\cdot\|$ be some norm for $\sdim\! \times\! \cdim$ matrices and $\|\cdot\|_{\star}$ its dual norm\footnote{$\|\Ub\|_{\star} \defin \sup_{\Db, \|\Db\| \leq 1} \langle \Ub,\Db\rangle_{F}$ with $\langle \cdot,\cdot\rangle_F$ the Frobenius inner product.}. 
For any $\epsilon > 0$, the set
\begin{align*}
	\Afrak_{\epsilon}(\Signal,\Db) 
	= \{&[\coeff_{1},\ldots,\coeff_{\nsamp}] \,:\, \coeff_i \in \R{d},\\
	&\Lcal_{\signal_{i}}(\Db,\coeff_{i}) \leq f_{\signal_{i}}(\Db)+\epsilon\}.
\end{align*}
is not empty, and for any $\Db'$ we have
\begin{equation*}
F_{\Signal}(\Db') \leq F_{\Signal}(\Db)  + L_{\Signal}(\Db)  \cdot \|\Db'-\Db\|+ C_{\Signal}(\Db) \cdot \|\Db'-\Db\|^{2}.
\end{equation*}
with
\begin{align}
\label{eq:DualNormResidualAssumption1}
L_{\Signal}(\Db) & \defin
\inf_{\epsilon>0} \sup_{\Coeff \in \Afrak_{\epsilon}} 
\tfrac{1}{\nsamp} \cdot \|(\Signal-\Db\Coeff)\Coeff^{\top}\|_{\star},\\
\label{eq:FroebeniusNormCoeffAssumption2}
C_{\Signal}(\Db)
& \defin
\inf_{\epsilon>0}\sup_{\Coeff \in \Afrak_{\epsilon}} 
\tfrac{C}{2\nsamp} \sum_{i=1}^{n }\|\coeff_{i}\|_{1}^{2}.
\end{align}
The constant $C$ in \eqref{eq:FroebeniusNormCoeffAssumption2} depends only on the norm $\|\cdot\|$ and the dimensions $\sdim, \cdim$. When $\|\cdot\| = \|\cdot\|_{1\to2}$ we have $C=1$. When $\Afrak_{0} \neq \emptyset$, $L_{\Signal}(\Db)$ and $C_{\Signal}(\Db)$ are suprema over $\Afrak_{0}$.
\end{lemma}
The choice of the $\ell^{1}$ norm in the bound~\eqref{eq:FroebeniusNormCoeffAssumption2} is an arbitrary convenience with no impact on the nature of the results to follow. More important will be the choice of the metric $\|\cdot\|$ which will be related to how we measure covering numbers for $\Cfrak$. This will be discussed later.

\begin{IEEEproof}
Fix $\epsilon > 0$ and $\Coeff \in \Afrak_{\epsilon}$. From the identity
\begin{align}
\label{eq:L_ineq}
 \Lcal_{\Signal}(\Db',\Coeff)  = {} & \Lcal_{\Signal}(\Db,\Coeff) + \tfrac{1}{\nsamp} \langle \Signal-\Db\Coeff, (\Db'-\Db)\Coeff \rangle_{F} \nonumber \\
 & + \tfrac{1}{2\nsamp}\|(\Db'-\Db)\Coeff\|_F^2
\end{align}
and the definition of the dual norm we have
\begin{align*}
\frac{| \langle \Signal-\Db\Coeff , (\Db'-\Db)\Coeff \rangle_{F}|}{\|\Db'-\Db\|}
&\leq
\left\| (\Signal-\Db\Coeff)\Coeff^{\top}\right\|_{\star}.
\end{align*}
Moreover, by the equivalence of all norms on the finite dimensional space of $\sdim \times \cdim$ matrices, there is a constant $C$ (equal to one when $\|\cdot\|=\|\cdot\|_{1\to 2}$) such that 
\begin{align*}
\tfrac{1}{2\nsamp} \|(\Db'-\Db)\Coeff\|_{F}^{2} 
&\leq \|\Db'-\Db\|_{1\to2}^{2} \cdot \tfrac{1}{2\nsamp} \sum_{i=1}^{n }\|\coeff_{i}\|_{1}^{2}
\\
&\leq \|\Db'-\Db\|^{2} \cdot \tfrac{C}{2\nsamp} \sum_{i=1}^{n }\|\coeff_{i}\|_{1}^{2}.
\end{align*}
Therefore, we deduce from~\eqref{eq:L_ineq} that 
\begin{align}
	F_{\Signal}(\Db') 
	= {} & \inf_{\Coeff} \Lcal_{\Signal}(\Db',\Coeff)
	\leq \sup_{\Coeff \in \Afrak_{\epsilon}} \Lcal_{\Signal}(\Db',\Coeff)\notag\\
	\leq {} & 
	\sup_{\Coeff \in \Afrak_{\epsilon}}  \Lcal_{\Signal}(\Db,\Coeff)\notag\\
	& + \sup_{\Coeff \in \Afrak_{\epsilon}} \tfrac{1}{\nsamp} \|(\Signal-\Db\Coeff)\Coeff^{\top}\|_{\star} \cdot \|\Db'-\Db\|\notag\\
	& + \sup_{\Coeff \in \Afrak_{\epsilon}} \tfrac{C}{2\nsamp} \sum_{i=1}^{n }\|\coeff_{i}\|_{1}^{2} \cdot \|\Db'-\Db\|^{2}\label{eq:fact1}
\end{align}
By definition, $\sup_{\Coeff \in \Afrak_{\epsilon}} \Lcal_{\Signal}(\Db,\Coeff) \leq F_{\Signal}(\Db) +\epsilon$. We conclude by taking the infimum of~\eqref{eq:fact1} over all $\epsilon>0$.
\end{IEEEproof}

\begin{corollary}\label{cor:LocalLipschitz}
Let $\|\cdot\|$ be some norm for $\sdim\! \times\! \cdim$ matrices and $\|\cdot\|_{\star}$ its dual norm, and $\Cfrak$ a class of dictionaries. 
If
\begin{align}
\label{eq:LConstantOverClass} \sup_{\Db \in \Cfrak} L_{\Signal}(\Db) & \leq L_{\Signal}(\Cfrak);\\
\label{eq:CConstantOverClass} \sup_{\Db \in \Cfrak} C_{\Signal}(\Db) & \leq C_{\Signal}(\Cfrak),
\end{align}
then for any 
$\Db \neq \Db' \in \Cfrak$ we have
\begin{equation*}
\frac{\left | F_{\Signal}(\Db')-F_{\Signal}(\Db) \right |}{\|\Db'-\Db\| } 
\leq L_{\Signal}(\Cfrak)  \cdot \left(1+\tfrac{C_{\Signal}(\Cfrak)}{L_{\Signal}(\Cfrak)} \|\Db'-\Db\|\right).
\end{equation*}
\end{corollary}
In particular, if we can establish the bounds~\eqref{eq:LConstantOverClass} and~\eqref{eq:CConstantOverClass} then $F_{\Signal}$ is uniformly locally Lipschitz (with respect to $\|\cdot\|$) over the class $\Cfrak$ for any constant $L > L_{\Signal}(\Cfrak)$.

\subsection{Uniform local Lipschitz constants}
Given Corollary~\ref{cor:LocalLipschitz} we now bound $L_{\Signal}(\Db)$ and $C_{\Signal}(\Db)$ 
when the norm $\|\cdot\|$ is the operator norm $\|\cdot\|_{1 \to 2}$. This is motivated by the fact that standard sparse coding is often performed with dictionaries constrained to have unit $\ell^{2}$ norm, which are closely connected with the unit sphere of this norm. With this choice, the shape of $\bar{g}$ in Definition~\ref{def:DefCoercivityFn} should appear natural in the context of the proof of Lemma~\ref{le:ConstantsWithNormPenalties_A1A3} below. 

Extensions to other definitions of $\bar{g}$ can be envisioned if we change the metric. In particular, when $g$ itself is a norm, one could consider the operator norm $\|\cdot\|_{g \to 2}$, at the price of possibly more tricky estimates of covering numbers.

\subsubsection{Penalty functions satisfying {\bf A1-A3}, with arbitrary $\Db$}
\begin{lemma}
\label{le:ArgMinSetNonEmpty}
If $g$ satisfies Assumptions {\bf A1-A3}, then the set $\Afrak_{0}$ is not empty, and it is bounded.
\end{lemma}
\begin{IEEEproof}
By the non-negativity ({\bf A1}) and coercivity ({\bf A3}) of $g(\cdot)$, $\Lcal_\Signal(\Db,\Coeff)$ is non-negative and $\lim_{k\to \infty} \Lcal_\Signal(\Db,\Coeff_k) = \infty$ whenever
$\lim_{k\to \infty}\|\Coeff_k\|=\infty$. Therefore, the function $\Coeff \mapsto \Lcal_{\Signal}(\Db,\Coeff)$ has bounded sublevel sets. Moreover, since $g$ is lower semi-continuous ({\bf A2}), then so is $\Coeff \mapsto \Lcal_\Signal(\Db,\Coeff)$, therefore it attains its infimum value. 
\end{IEEEproof}

\begin{lemma}[Penalty functions that satisfy~{\bf A1-A3}]
\label{le:ConstantsWithNormPenalties_A1A3}
Under Assumptions {\bf A1-A3}, for any training set $\Signal$ and dictionary $\Db$, the constants defined in~\eqref{eq:DualNormResidualAssumption1} and~\eqref{eq:FroebeniusNormCoeffAssumption2} with the norm $\|\cdot\| = \|\cdot\|_{1 \to 2}$ satisfy the bounds
\begin{align}
\label{eq:ConstantsWithNormPenalties1_A1A3}
	L_{\Signal}(\Db) 
	& \leq 
	\tfrac{1}{\nsamp} \sum_{i=1}^{\nsamp} \sqrt{ 2f_{\signal_{i}}(\Db)} \cdot \bar{g}(f_{\signal_{i}}(\Db)) \\
	\label{eq:ConstantsWithNormPenalties_A1A3}
	C_{\Signal}(\Db) 
	& \leq 
	\tfrac{1}{2\nsamp} \sum_{i=1}^{\nsamp} [\bar{g}(f_{\signal_{i}}(\Db))]^{2}.
\end{align}
\end{lemma}

\begin{IEEEproof}
By Lemma~\ref{le:ArgMinSetNonEmpty} the set $\Afrak_{0}$ is not empty, and by definition, for any $\Coeff \defin [\coeff_{1}, \ldots, \coeff_{\nsamp}] \in \Afrak_{0}$,
\begin{equation}
\label{eq:NearBestCoeff}
	\tfrac{1}{2}\| \signal_{i}-\Db \coeff_{i}\|_{2}^{2} + g(\coeff_{i}) 
 \leq f_{\signal_{i}}(\Db).
\end{equation}
Therefore, {\bf A1} (non-negativity of $g$) implies
\begin{align}
	\label{eq:gBoundCrude_A1A3}
	0 \leq g(\coeff_{i}) & \leq f_{\signal_{i}}(\Db)\\
	\label{eq:residualBoundCrude_A1A3}
	\|\signal_{i}-\Db\coeff_{i}\|_{2} & \leq  \sqrt{2 f_{\signal_{i}}(\Db)},
\end{align}
for $i=1,\ldots,\nsamp$. Combined with Definition~\ref{def:DefCoercivityFn},~\eqref{eq:gBoundCrude_A1A3} implies
\begin{align}
	\|\coeff_{i}\|_{1} & \leq \bar{g}(f_{\signal_{i}}(\Db))\label{eq:coeffBound2_A1A3},\\
	C_{\Signal}(\Db) & \leq \tfrac{1}{2\nsamp}\sum_{i=1}^{\nsamp}[\bar{g}(f_{\signal_{i}}(\Db))]^{2},\label{eq:Cbound_A1A3}
\end{align}
which proves the bound in~\eqref{eq:ConstantsWithNormPenalties_A1A3}.
We now prove the inequality~\eqref{eq:ConstantsWithNormPenalties1_A1A3}. For any $\sdim\! \times\! \cdim$ matrix $\Deltab$ we have
\begin{align}
	\nonumber
	\langle \Signal-\Db \Coeff, \Deltab\Coeff\rangle_{F} 
	&=\sum_{i=1}^{\nsamp} \langle \signal_{i}-\Db\coeff_{i},\Deltab \coeff_{i}\rangle\\
	\nonumber
	&\leq \sum_{i=1}^{\nsamp} \| \signal_{i}-\Db\coeff_{i}\|_{2} \cdot \|\Deltab \coeff_{i}\|_{2}\\
	\label{eq:Lip3}&\leq \sum_{i=1}^{\nsamp} \| \signal_{i}-\Db\coeff_{i}\|_{2} \cdot \|\Deltab\|_{1 \to 2} \cdot \|\coeff_{i}\|_{1}.\nonumber\\
\end{align}
This allows us to provide an upper bound for the dual norm by exploiting Equations~\eqref{eq:residualBoundCrude_A1A3}  and~\eqref{eq:coeffBound2_A1A3}
\begin{equation}
\label{eq:LBound_A1A3}
	\tfrac{1}{\nsamp} \|(\Signal-\Db\Coeff)\Coeff^{\top}\|_\star 
	\leq 
	\tfrac{1}{\nsamp} \sum_{i=1}^{\nsamp} \sqrt{ 2 f_{\signal_{i}}(\Db)} \cdot \bar{g}(f_{\signal_{i}}(\Db))
\end{equation}
which shows \eqref{eq:ConstantsWithNormPenalties1_A1A3}.
\end{IEEEproof}

\begin{corollary}[Penalty functions that satisfy~{\bf A1-A4}]
\label{le:ConstantsWithNormPenalties}
Under Assumptions {\bf A1-A4}, for any training set $\Signal$ and dictionary $\Db$, the constants defined in~\eqref{eq:DualNormResidualAssumption1} and~\eqref{eq:FroebeniusNormCoeffAssumption2} with the norm $\|\cdot\| = \|\cdot\|_{1 \to 2}$ satisfy the bounds
\begin{align}
\label{eq:ConstantsWithNormPenalties1}
	L_{\Signal}(\Db) 
	{}&\leq \tfrac{1}{\nsamp} \sum_{i=1}^{\nsamp} \|\signal_{i}\|_2 \cdot \bar{g}(\tfrac{1}{2}\|\signal_{i}\|_2^2) 
	&&\defin L_{\Signal}(\bar{g})\\
	\label{eq:ConstantsWithNormPenalties}
	C_{\Signal}(\Db) 
	& \leq 
	\tfrac{1}{2\nsamp} \sum_{i=1}^{\nsamp} [\bar{g}(\|\signal_{i}\|_{2}^{2}/2)]^{2} 
	&&\defin 
	C_{\Signal}(\bar{g}).
\end{align}
\end{corollary}

\begin{IEEEproof}
If the condition \textbf{A4} ($g(\mathbf{0})=0$) is fulfilled, we can extend Equations~(\ref{eq:gBoundCrude_A1A3},\ref{eq:residualBoundCrude_A1A3}) to
\begin{align}
	\label{eq:gBoundCrude}
	0 \leq g(\coeff_{i}) & \leq f_{\signal_{i}}(\Db)  \leq \Lcal_{\signal_{i}}(\Db,\mathbf{0}) = \tfrac{1}{2} \|\signal_{i}\|_{2}^{2}\\
	\label{eq:residualBoundCrude}
	\|\signal_{i}-\Db\coeff_{i}\|_{2} & \leq  \sqrt{2 f_{\signal_{i}}(\Db)} \leq \|\signal_{i}\|_{2},
\end{align}
for $i=1,\ldots,n$. Due to the fact that $\bar{g}$ is non-decreasing, the inequality in~\eqref{eq:ConstantsWithNormPenalties} follows directly from Equation~\eqref{eq:Cbound_A1A3}.

It remains to show the inequality in \eqref{eq:ConstantsWithNormPenalties1}, which is verified by plugging Equations~(\ref{eq:gBoundCrude},\ref{eq:residualBoundCrude}) in \eqref{eq:LBound_A1A3}.
This yields the upper bound 
\begin{align*}
	\tfrac{1}{\nsamp} \sum_{i=1}^{\nsamp} \sqrt{ 2 f_{\signal_{i}}(\Db)} \cdot \bar{g}(f_{\signal_{i}}(\Db)) \nonumber
	{} \leq \tfrac{1}{\nsamp} \sum_{i=1}^{\nsamp} \|\signal_{i}\|_2 \cdot \bar{g}(\tfrac{1}{2}\|\signal_{i}\|_2^2) 
\end{align*}
which is independent of the constraint set $\Cfrak$.
\end{IEEEproof}

\subsubsection{Penalty functions and constraint sets that satisfy {\bf B1-B3}}

\begin{lemma}[Penalty functions $g$ and constraint sets $\Cfrak$ that satisfy {\bf B1-B3}]
\label{le:ConstantsWithKSparse}
Under assumptions  {\bf B1-B3}, for any training set $\Signal$ and dictionary $\Db \in \Cfrak$, the constants defined in~\eqref{eq:DualNormResidualAssumption1} and~\eqref{eq:FroebeniusNormCoeffAssumption2} with the norm $\|\cdot\| = \|\cdot\|_{1 \to 2}$ satisfy the bounds%
\begin{align}
	\label{eq:ConstantsWithKSparse1}
	L_{\Signal}(\Db)
	& \leq
	\tfrac{2}{\nsamp\sqrt{\kappa}} \|\Signal\|_{F}^{2} && \defin L_{\Signal}(\bar{g})\\
	\label{eq:ConstantsWithKSparse2}
	C_{\Signal}(\Db) 
	&\leq
	\tfrac{2}{\nsamp \kappa} \|\Signal\|_{F}^{2} && \defin C_{\Signal}(\bar{g}).
\end{align}
\end{lemma}

\begin{IEEEproof} 
Fix $\epsilon>0$ and consider $\coeff_i$, $i=1,\ldots,\nsamp$ such that $\Coeff \defin [\coeff_1,\ldots,\coeff_\nsamp] \in \Afrak_{\epsilon}$. One can check that by {\bf B1}, Equations~\eqref{eq:NearBestCoeff}-\eqref{eq:residualBoundCrude_A1A3} hold up to an additive $O(\epsilon)$ term\footnote{$f(\epsilon) = O(\epsilon)$ if there exists $C<\infty$ such that $|f(\epsilon)| \leq C\epsilon$.} on the right hand side, as well as~\eqref{eq:Lip3}. By {\bf B3} we also have~\eqref{eq:gBoundCrude}-\eqref{eq:residualBoundCrude} with a similar additive term, yielding
\begin{equation}
	\|\Db\coeff_i\|_2 \leq \|\signal_i\|_2 + \|\signal_i - \Db \coeff_i\| \leq 2 \| \signal_i \| + O(\epsilon).
\end{equation}
Moreover, by {\bf B2}, we have
\begin{equation}
\label{eq:IntermediateEstimateCoeffKSparse}
\|\coeff_{i}\|_{1}^{2} \leq \tfrac{\|\Db\coeff_i\|_2^2}{\kappa} \leq \frac{4 \| \signal_i \|_2^2}{\kappa} + O(\epsilon).
\end{equation}
Taking the infimum over $\epsilon>0$ shows~\eqref{eq:ConstantsWithKSparse2}. 

To prove~\eqref{eq:ConstantsWithKSparse1} we combine~\eqref{eq:Lip3} with~\eqref{eq:residualBoundCrude} and~\eqref{eq:IntermediateEstimateCoeffKSparse}, yielding an upper bound to $\tfrac{1}{n} \|(\Signal-\Db\Coeff)\Coeff^{\top}\|_{\star}$:
\begin{equation}
	\tfrac{1}{\nsamp} \sum_{i=1}^\nsamp \|\signal_i - \Db \coeff_i\|_2 \cdot \|\coeff_i\|_1 \leq \frac{2}{\nsamp\sqrt{\kappa}} \sum_{i=1}^\nsamp \|\signal_i \|_2^2 + O(\epsilon).
\end{equation}
\end{IEEEproof}

Observe the similarity between the expressions~\eqref{eq:ConstantsWithKSparse1} and~\eqref{eq:DefLipschitzConstantUniform}. It justifies Definition~\ref{def:DefBarGCaseB} which yields the identity $L_{\Signal}(\Cfrak) = L_{\Signal}(\bar{g})$. With this notation we will be able to jointly cover the different penalty functions dealt with so far.

\subsection{Main results on Lipschitz properties}
\label{sec:mainLipschitz}
We are ready to state the main results on the Lipschitz property of $F_{\Signal}(\Db)$. First we go for a {\em global} Lipschitz property. Note that Theorem~\ref{th:LipschitzNormPenalties} below  is expressed over a {\em convex} class of dictionaries: under assumptions {\bf A1-A4} the result is valid uniformly over $\Db \in \R{\sdim \times \cdim}$ (in particular, {\em it is not restricted to dictionaries with unit norm columns})\footnote{This requirement will only arise from the need to have a finite covering number for the sample complexity estimate}; under assumptions {\bf B1-B3}, we add the explicit assumption {\bf B4} that $\Cfrak$ is convex.

\begin{theorem}\label{th:LipschitzNormPenalties}
Assume either {\bf A1-A4} or {\bf B1-B4}. Then, for any $\Signal$, and any $\Db,\Db' \in \Cfrak$, 
\begin{equation}
\label{eq:LipschitzWithNormPenalties}
	\left | F_{\Signal}(\Db)-F_{\Signal}(\Db') \right | \leq L_{\Signal}(\bar{g})  \cdot \|\Db-\Db'\|_{1 \to 2}.
\end{equation}
\end{theorem}

\begin{IEEEproof}
Fix $\epsilon>0$. By Corollaries~\ref{cor:LocalLipschitz} and~\ref{le:ConstantsWithNormPenalties} (resp. Lemma~\ref{le:ConstantsWithKSparse}):
\begin{equation}
\label{eq:LocalLipschitz}
	\left| F_{\Signal}(\Db') - F_{\Signal}(\Db)\right|  \leq  (1+\epsilon)  L_{\Signal}(\bar{g}) \cdot \|\Db'-\Db\|_{1\to 2}
\end{equation}
whenever $\delta = \|\Db'-\Db\|_{1\to2} \leq \tfrac{\epsilon L_{\Signal}(\bar{g})}{C_{\Signal}(\bar{g})}$.

When $\delta$ exceeds this bound, we choose \RGrev{an integer} $k \geq 1$ such that $\delta/k  \leq \tfrac{\RGrev{\epsilon} L_{\Signal}(\bar{g})}{C_{\Signal}(\bar{g})}$ and define $\Db_{i} = \Db + i (\Db'-\Db)/k$, $0\leq i \leq k$. 
Note that this sequence is a priori constructed in the surrounding space of $\Cfrak$. Under assumptions {\bf B1-B4}, the convexity of $\Cfrak$ ({\bf B4}) ensures $\Db_{i} \in \Cfrak$. Under assumptions {\bf A1-A4}, the local Lipschitz constant $L_\Signal(\bar{g})$ defined in \eqref{eq:ConstantsWithNormPenalties1} actually holds independently of a particular dictionary structure $\Cfrak$. Hence, in both cases, since $\|\Db_{i+1}-\Db_{i}\|_{1 \to 2} \leq \tfrac{\RGrev{\epsilon} L_{\Signal}(\bar{g})}{C_{\Signal}(\bar{g})}$ for $i=\RGrev{0},\ldots,k-1$, the bound we just obtained yields:
\begin{eqnarray*}
\left| F_{\Signal}(\Db_{i+1}) - F_{\Signal}(\Db_{i})\right| 
\leq & (1+\epsilon) \  L_{\Signal}(\bar{g}) \ \|\Db_{i+1} - \Db_{i}\|_{1\to2} \\
 \leq & (1+\epsilon) \ L_{\Signal}(\bar{g}) \ \|\Db' - \Db\|_{1\to2}/k\\
\left| F_{\Signal}(\Db') - F_{\Signal}(\Db)\right| 
\leq& \sum_{i=0}^{k-1} |F_{\Signal}(\Db_{i+1}) - F_{\Signal}(\Db_{i})|\\
 \leq& (1+\epsilon)\  L_{\Signal}(\bar{g})\  \|\Db' - \Db\|_{1\to2}.
\end{eqnarray*}
Thus, the bound~\eqref{eq:LocalLipschitz} can be extended to any pair $\Db,\Db' \in \Cfrak$. Since the choice of $\epsilon>0$ is arbitrary, Equation~\eqref{eq:LipschitzWithNormPenalties} follows.
\end{IEEEproof}

As an example, consider NMF expressed with $g = \chiNMF$ and $\Cfrak_{\textrm{NMF} \& \ell^{1}}$ the class of non-negative dictionaries with $\ell_{1}$ normalized columns, used in topic models \cite{Jenatton:2011vv},
\begin{equation*}\begin{split}
	\Cfrak_{\textrm{NMF} \& \ell^{1}} \defin \{ \Db \,:\, \db_i \in\R{\sdim}; \|\db_{i}\|_{1} = 1;\, d_{ij} \geq 0\},
\end{split}\end{equation*}
which is convex ({\bf B4}) as it is the Cartesian product of $\cdim$ copies of the simplex. Since $\Cfrak \subset \CNMF(\kappa)$ with $\kappa = 1/\sdim\cdim$, assumptions {\bf B1-B3} hold.

A slightly weaker result expressed in terms of {\em uniform local} Lipschitz property holds under {\bf B1-B3} for {\em non-convex} $\Cfrak$.

\begin{theorem}\label{th:LipschitzKSparse}
Under assumptions {\bf B1-B3},
for any training set $\Signal$ and any dictionaries $\Db' \neq \Db \in \Cfrak$
we have
\begin{equation*}
	\frac{\left | F_{\Signal}(\Db')-F_{\Signal}(\Db) \right |}{\|\Db'-\Db\|_{1 \to 2}}
	\leq L_{\Signal}(\bar{g}) \cdot \left(1+ \sqrt{\tfrac{1}{\kappa}} \|\Db'-\Db\|_{1 \to 2}\right).
\end{equation*}
\end{theorem}

\begin{IEEEproof}
This is the direct result of Lemma~\ref{le:OneSidedLipschitz}. The upper bounds for $C_\Signal(\Cfrak), L_\Signal(\Cfrak)$ provided in Lemma~\ref{le:ConstantsWithKSparse} yield the factor $\tfrac{C_{\Signal}(\bar{g})}{L_{\Signal}(\bar{g})} = \sqrt{\tfrac{1}{\kappa}}$.
\end{IEEEproof}

\begin{remark}
The reader may wonder why we state a global Lipschitz property in Theorem~\ref{th:LipschitzNormPenalties} but only a uniform local one in Theorem~\ref{th:LipschitzKSparse}. First, the triangle inequality argument does not seem to extend: the line joining $\Db$ to $\Db'$ cannot be cut in small segments so that each intermediate pair remains in $\Cfrak$.
In fact, there may even be several connected components in the class. Moreover, even in a given connected component, it is not clear what the length of the geodesics that would mimic such a triangle inequality argument is. 
Answering such questions would require a deeper investigation of the topology and geometry of $\Cfrak$.  
Since Theorem~\ref{th:LipschitzKSparse} is sufficient for our needs, we leave them to future work. 
\end{remark}

\FBrev{
\subsection{Lipschitz-continuity for high-dimensional settings}
\label{sec:Lipschitz_alt}
When the signal dimension $\sdim$ is larger than the number $\cdim$ of dictionary elements, then the Lipschitz-continuity studied above is not appropriate as it leads to sample complexities that grow with $\sdim$. Instead, we may consider Lipschitz-continuity with respect to $\Db^{\top}\Db$ and $\Db^\top \Signal$, which we show below and use in Section~\ref{sec:sample_alt} to sketch improved results in such high-dimensional settings.

\begin{lemma}\label{le:OneSidedLipschitz_alt}
Under the same assumptions as in Lemma~\ref{le:OneSidedLipschitz}, for any $\Db'$, we have
\begin{equation*}
F_{\Signal}(\Db') \leq F_{\Signal}(\Db)  +
C^{(1)}_{\Signal}(\Db) \cdot
 \|(\Db')^\top \Db' -\Db^\top \Db \|_F+ 
 C^{(2)}_{\Signal}(\Db) \cdot \|\Signal^\top \Db'-\Signal^\top \Db\|_F.
\end{equation*}
with
\begin{align*}
C^{(1)}_{\Signal}(\Db) & \defin
\inf_{\epsilon>0} \sup_{\Coeff \in \Afrak_{\epsilon}} 
\tfrac{1}{2 \nsamp} \cdot \|\Coeff \Coeff^\top\|_{F},\\
C^{(2)}_{\Signal}(\Db) & \defin
\inf_{\epsilon>0} \sup_{\Coeff \in \Afrak_{\epsilon}} 
\tfrac{1}{ \nsamp} \cdot \|\Coeff \|_{F}.
 \end{align*}
\end{lemma}
\begin{IEEEproof}
We follow the same principle as in the proof of Lemma~\ref{le:OneSidedLipschitz}, but we use the following equality
\begin{align*}
 \Lcal_{\Signal}(\Db',\Coeff)  = {} & \Lcal_{\Signal}(\Db,\Coeff) + \tfrac{1}{2\nsamp} 
 \langle 
 (\Db')^\top \Db' -\Db^\top \Db, \Coeff \Coeff^\top
 \rangle_{F} - \tfrac{1}{ \nsamp}  \langle 
(\Db')^{\top}\Signal-\Db^{\top}\Signal, \Coeff 
 \rangle_{F}.
\end{align*}
Note that we use the Frobenius norm for this lemma, but that we could use any norm on matrices.
\end{IEEEproof}
}

\RGrev{
\subsection{Other losses}\label{sec:other losses}
One can envision extensions of Lemmata~\ref{le:OneSidedLipschitz} and~\ref{le:OneSidedLipschitz_alt} and their consequences to matrix factorization problems where the $\ell_{2}$ data-fidelity term, $\tfrac{1}{2}\|\signal-\Db\coeff\|_{2}^{2}$, is replaced by a more general loss $\ell(\signal,\Db\coeff)$. 
In particular, Lemma~\ref{le:OneSidedLipschitz} can be extended to losses $\ell(\cdot,\cdot)$ satisfying for any $\mathbf{x},\mathbf{y}$
\begin{equation*}
\ell(\signal,\mathbf{y}+\mathbf{h}) \leq \ell(\signal,\mathbf{y}) + \langle \nabla_{\mathbf{y}} \ell(\signal,\mathbf{y}), \mathbf{h}\rangle + C \|\mathbf{h}\|_{2}^{2},
\end{equation*}
i.e., with a local quadratic behavior. This would be reminiscent of the work of Negahban {\em et al} \cite{Negahban:2012en} which covers M-estimation problems where the loss, which is convex in the unknown parameter is regularized with a convex decomposable penalty. Here the considered problem is intrinsically non convex in $\Db$. 

A generalization of Corollary~\ref{le:ConstantsWithNormPenalties} would further require assuming that $\ell(\signal,\mathbf{y}) \geq 0$ for any $\mathbf{x},\mathbf{y}$, to replace~\eqref{eq:gBoundCrude} with $0 \leq g(\coeff_{i}) \leq \ell(\signal,\mathbf{0})$, and identifying conditions on the loss ensuring that for some function $B(.)$, an analogue of the bound~\eqref{eq:residualBoundCrude} holds: $\|\nabla_{\mathbf{y}} \ell(\signal_{i},\Db\coeff_{i})\|_{2} \leq B(\signal_{i})$. The resulting Lipschitz constant would read $L_{\Signal} = \tfrac{1}{\nsamp} \sum_{i=1}^{\nsamp} B(\signal_{i}) \bar{g}(\ell(\signal_{i},\mathbf{0}))$. The full characterization of such extensions and of the families of losses that can be considered is, however, beyond the scope of this paper.  
}


\section{Sample Complexity (Proof of Theorem~\ref{th:MainTheorem})}
\label{sec:Concentration}

Given the global (resp. uniform local) Lipschitz property of $F_{\Signal}$, a standard route (see, e.g., \cite{MaurerPontil,vainsencher:2010}) to control the sample complexity via a uniform convergence result is to rely on concentration of measure and covering numbers for the considered class of dictionaries (see Section~\ref{sec:Covering} for details on covering numbers).

\subsection{Lipschitz property of the expected cost function}
Mild assumptions on the quantities $\Lipdev(L)$ and $\Expdev(\gamma)$ as defined in \eqref{eq:DefProbLipschitz} and \eqref{eq:DefProbConcentration} are sufficient to control the Lipschitz constants of $F_{\Signal}$ and of its expectation: 
\\
\begin{lemma}
\label{lm:probabilistic_control}
Under assumptions {\bf A1-A4} or {\bf B1-B4}, and {\bf C1}:
 \begin{enumerate}
\item  
the function $\Db \to F_{\Signal}(\Db)$ is Lipschitz with constant $L$ , with probability at least $1-\Lipdev(L)$. 
\item the expected cost function $\Db \mapsto \EE f_{\signal}(\Db)$ is Lipschitz with constant $L$ as soon as $L > L_{\PP}(\bar{g})$ and:
\begin{itemize}
\item[{\bf C2'}] there exists a sequence $\gamma_{\nsamp}$ such that
\begin{align*}
\lim_{\nsamp \to \infty} \gamma_{\nsamp} &= 0; &
\lim_{\nsamp \to \infty} \Expdev(\gamma_{\nsamp}) &= 0
\end{align*}
\end{itemize}
\end{enumerate}
\end{lemma}
\begin{IEEEproof}
The first result trivially follows from Section~\ref{sec:Lipschitz} and the definition of $\Lipdev(L)$. For the second one, given $\Db,\Db'$, consider an i.i.d.\ draw $\Signal$ of $\nsamp$ samples from $\PP$. We have
\begin{align*}
|\EE f_{\signal}(\Db')-\EE f_{\signal}(\Db)|
\leq {} &
|\EE f_{\signal}(\Db')-F_{\Signal}(\Db')|\\
& + |F_{\Signal}(\Db')-F_{\Signal}(\Db)| \\
& + |F_{\Signal}(\Db)-\EE f_{\signal}(\Db)|\\
\leq {} & L \|\Db'-\Db\|_{1\to2} + 2 \gamma_{\nsamp}
\end{align*}
except with probability at most $\Lipdev(L) + 2 \Expdev(\gamma_{\nsamp})$. The limit for large $n$ yields the desired bound with probability one.
\end{IEEEproof}

\begin{remark}
Under {\bf B1-B3} only (without the convexity assumption on $\Cfrak$), 
an analogon of Lemma~\ref{lm:probabilistic_control} holds where the conclusion that $\Db \mapsto F_{\Signal}(\Db)$ (resp. its expectation)  is ``Lipschitz with constant $L$'' is replaced with:
\begin{equation}
\label{eq:QuasiLipschitz}
\frac{\left | F_{\Signal}(\Db')-F_{\Signal}(\Db) \right |}{\|\Db-\Db'\|_{1 \to 2}}
\leq L \cdot \left(1+ \sqrt{\tfrac{1}{\kappa}} \|\Db-\Db'\|_{1 \to 2}\right).
\end{equation}
(resp.~\eqref{eq:QuasiLipschitz} holds with $F_{\Signal}(\cdot)$ replaced by $\EE f_{\signal}(\cdot)$).
\end{remark}

We are now ready to state a first uniform convergence result. 

\subsection{Abstract uniform convergence result}

\begin{lemma}\label{le:CoveringArgumentGeneric}
Assume {\bf C1-C2}. For any $\epsilon,\gamma>0$ we have, except with probability at most 
\begin{equation}
\label{eq:ProbaGeneric}
\Lipdev(L) + \Ncal(\Cfrak,\epsilon) \cdot \Expdev(\gamma)
\end{equation}
\begin{itemize}
\item under  {\bf A1-A4} or {\bf B1-B4}\textup{:}
\begin{equation}
\label{eq:UniformConvergence_result1}
	\sup_{\Db \in \Cfrak} \left| F_{\Signal}(\Db)-\EE f_{\signal}(\Db)\right| 
	\leq 2L\epsilon + \gamma
\end{equation}
\item under {\bf B1-B3} only\textup{:}
\begin{equation}
\label{eq:UniformConvergence_result2}
	\sup_{\Db \in \Cfrak} \left| F_{\Signal}(\Db)-\EE f_{\signal}(\Db)\right| 
	\leq 2L\epsilon \cdot \left(1+ \sqrt{\tfrac{1}{\kappa}} \epsilon\right) + \gamma
\end{equation}
\end{itemize}
\end{lemma}
\begin{IEEEproof}
\color{black}
We only give the proof under \textbf{A1-A4} or \textbf{B1-B4}. A straightforward adaptation yields the proof under \textbf{B1-B3} only.

Fix $\epsilon>0$ and consider an $\epsilon$-cover of $\Cfrak$ with respect to the $\|\cdot\|_{1 \to 2}$ metric with at most $\Ncal = \Ncal(\Cfrak,\epsilon)$ elements $\{\Db_{j}\}$. Fixing $\Db \in \Cfrak$ there is an index $j$ such that $\|\Db_{j} -\Db\|_{1\to2} \leq \epsilon$, hence
\begin{align*}
\left| F_{\Signal}(\Db)-\EE f_{\signal}(\Db)\right| 
\leq {} &
\left| F_{\Signal}(\Db)-F_{\Signal}(\Db_{j})\right| \\
& +\left| F_{\Signal}(\Db_{j})-\EE f_{\signal}(\Db_{j})\right| \\
& +\left| \EE f_{\signal}(\Db_{j})-\EE f_{\signal}(\Db)\right|\\
\leq {} &
\left| F_{\Signal}(\Db)-F_{\Signal}(\Db_{j})\right| \\
& +\sup_{1 \leq j \leq \Ncal}\left| F_{\Signal}(\Db_{j})-\EE f_{\signal}(\Db_{j})\right| \\
& + L\epsilon
\end{align*}
where we used the Lipschitz property of the expected cost function (Lemma~\ref{lm:probabilistic_control}). By a union bound, for any $\gamma>0$ we obtain
the bound~\eqref{eq:UniformConvergence_result1}
except with probability at most~\eqref{eq:ProbaGeneric}.
\end{IEEEproof}

\subsection{Main uniform convergence result}

The above sample complexity estimate is not quite explicit since it  
combines covering numbers and concentration bounds. 
In the following, we give more explicit sample complexity estimates under more specific assumptions on $\Ncal(\Cfrak,\epsilon)$ and $\Expdev(\gamma)$. 
The assumption on $\Ncal(\Cfrak,\epsilon)$ is justified by the fact that all examples of classes $\Cfrak$ developed in Section~\ref{sec:Covering} have covering numbers with bounds of the type expressed in~\eqref{eq:ExplicitGenericCoveringAssumption}. The assumption {\bf C2} on $\Expdev(\gamma)$ is further discussed at the end of this section.

\begin{lemma}\label{le:ExplicitSampleComplexity}
Assume {\bf C1-C2} holds and that $\Ncal(\Cfrak,\epsilon)$ satisfy the bound~\eqref{eq:ExplicitGenericCoveringAssumptionTh}, i.e., for $c>0$, $T \in (0,\infty]$, and $C,\cvdim\geq 1$\textup{:}
\begin{align}
\label{eq:ExplicitConcentrationAssumptionTh}
\Expdev(c\tau) & \leq 2 \exp(-\nsamp \tau^{2}),\ \forall 0 \leq \tau \leq T;\\
\label{eq:ExplicitGenericCoveringAssumptionTh}
\Ncal(\Cfrak,\epsilon) & \leq \left(\frac{C}{\epsilon}\right)^{\cvdim},\ \forall 0<\epsilon\leq 1.
\end{align}
Define 
\begin{equation*}
\beta \defin \cvdim \cdot \max(\log \tfrac{2LC}{c},1)
\end{equation*}
and $D=1$ (under \textbf{A1-A4} or \textbf{B1-B4}) or $D = \max(1/\kappa,1)$ (under \textbf{B1-B3}). 
Assume that the sample size $\nsamp$ satisfies
\begin{equation}
\label{eq:BasicSampleComplexity}
	\frac{\nsamp}{\log \nsamp} \geq \max\left(8,\tfrac{\beta}{T^{2}},D \cdot \left(\tfrac{c}{2L}\right)^{2}\beta\right).
\end{equation}
Then, for any
\begin{equation}
\label{eq:Hyp1}
0\leq  x \leq \nsamp T^{2}-\beta \log n
\end{equation}
we have, except with probability at most $\Lipdev(L) + 2 e^{-x}$
\begin{equation*}
	\sup_{\Db \in \Cfrak} \left| F_{\Signal}(\Db)-\EE f_{\signal}(\Db)\right| 
	\leq \eta_{\nsamp}(L,C,\cvdim,c).
\end{equation*}
In the case of {\bf A1-A4} or {\bf B1-B4}, we have
\begin{align*}
	\eta_{\nsamp} &\defin  2c \sqrt{\tfrac{\beta \log \nsamp}{\nsamp}} + c \sqrt{ \tfrac{\beta+x}{\nsamp}},\\
\intertext{whereas for {\bf B1-B3}, we obtain} 
	\eta_{\nsamp} &\defin  3c \sqrt{\tfrac{\beta \log \nsamp}{\nsamp}} + c \sqrt{ \tfrac{\beta+x}{\nsamp}}.
\end{align*}
\end{lemma}
Note that Theorem~\ref{th:MainTheorem} follows from Lemma~\ref{le:ExplicitSampleComplexity}.

\begin{IEEEproof}
First we observe that the condition \textbf{C2} on $\Expdev$ implies \textbf{C2'} (defined in Lemma~\ref{lm:probabilistic_control}), and since we assume \textbf{C1} we can apply Lemma~\ref{le:CoveringArgumentGeneric}.
Notice that~\eqref{eq:BasicSampleComplexity} implies that one can indeed find $x$ that satisfies assumption~\eqref{eq:Hyp1}. 
We set
\begin{align*}
\epsilon &= \tfrac{c\sqrt{\beta}}{2L} \sqrt{\tfrac{\log \nsamp}{\nsamp}},\\
\tau &= \sqrt{\tfrac{ \cvdim [\log C/\epsilon] + x}{\nsamp}} = \sqrt{\tfrac{\cvdim [\log (2LC/c\sqrt{\beta})]+ \tfrac{\cvdim}{2} \log \tfrac{\nsamp}{\log \nsamp}+x}{\nsamp}}.
\end{align*}
Since~\eqref{eq:BasicSampleComplexity} implies $\nsamp\geq 8$, hence $\tfrac{\log n}{2} \geq 1$, we have
\begin{equation*}
0\leq  x \leq \nsamp T^{2}-\tfrac{\beta}{2}\log \nsamp -\tfrac{\beta}{2} \log \nsamp \leq \nsamp T^{2}-\tfrac{\cvdim}{2} \log \nsamp -\beta.
\end{equation*}
By definition of $\beta$ and $\cvdim$ we have $\beta \geq \cvdim \geq 1$ hence
\begin{equation*}\begin{split}
\nsamp \tau^{2}- \tfrac{\cvdim}{2} \log \tfrac{\nsamp}{\log \nsamp}
& = \cvdim \log \tfrac{2LC}{c\sqrt{\beta}}+x
\leq 
 \cvdim \log \tfrac{2LC}{c} + x\\
{}&\leq
\beta +x
\leq \nsamp T^{2}-\tfrac{\cvdim}{2} \log \tfrac{\nsamp}{\log \nsamp}.
\end{split}\end{equation*}
This shows that $0\leq \tau \leq T$. Moreover, by~\eqref{eq:BasicSampleComplexity} we further have $0<\epsilon \leq 1$, hence we can apply~\eqref{eq:ExplicitConcentrationAssumptionTh} and~\eqref{eq:ExplicitGenericCoveringAssumptionTh} to obtain
\begin{equation*}
\Ncal(\Cfrak,\epsilon) \cdot \Expdev(c\tau) \leq 2 \cdot (C/\epsilon)^{\cvdim} \cdot \exp(-\nsamp\tau^{2}) = 2 \cdot e^{-x}.
\end{equation*}
Using \textbf{A1-A4} or \textbf{B1-B4} we conclude as follows: since $\beta \geq 1$ and $\log \nsamp \geq 1$ we have
\begin{align*}
2L\epsilon + c\tau
= {} &
c \sqrt{\tfrac{\beta \log \nsamp}{\nsamp}} \\
& + c \cdot \sqrt{
\cvdim \log \tfrac{2LC}{c\sqrt{\beta}} + \tfrac{\cvdim}{2} \log \tfrac{\nsamp}{\log \nsamp}+x}\cdot \tfrac{1}{\sqrt{\nsamp}}\\
\leq {} &
c \sqrt{\tfrac{\beta \log \nsamp}{\nsamp}}
+ c \cdot \sqrt{
\tfrac{\beta}{2} \log \nsamp + \beta+x } \cdot \tfrac{1}{\sqrt{\nsamp}}\\
\leq {} &
c \sqrt{\tfrac{\beta \log \nsamp}{\nsamp}} \cdot (1+\tfrac{1}{\sqrt{2}})
+ c \cdot \sqrt{\beta +x} \cdot \sqrt{\tfrac{1}{\nsamp}}\\
\leq {} &
2c  \sqrt{\tfrac{\beta \log \nsamp}{\nsamp}} + c \sqrt{\tfrac{\beta+x}{\nsamp}}.
\end{align*}
Under \textbf{B1-B3},
the definition of $D$ and assumption~\eqref{eq:BasicSampleComplexity} imply $0< \sqrt{\tfrac{1}{\kappa}}\epsilon \leq 1$ hence we get similarly:
\begin{equation*}
2L\epsilon \cdot \left(1+\sqrt{\tfrac{1}{\kappa}}\epsilon\right) + c\tau
\leq
3c  \sqrt{\tfrac{\beta \log \nsamp}{\nsamp}} + c\sqrt{\tfrac{\beta+x}{\nsamp}}.
\end{equation*}
\end{IEEEproof}

\subsection{On assumptions \textbf{C1} and \textbf{C2}}
Assumptions \textbf{C1-C2} are actually satisfied under rather standard hypotheses:

\begin{lemma}\label{le:ConcentrationSphere}
Assume that $\PP \in \Pfrak_{\radius\Sphere^{\sdim-1}}$. Then 
\begin{align}
\label{eq:ConcentrationSphere}
	\Expdev(\radius^{2}\tau/\sqrt{8}) &\leq 2 \exp(-\nsamp\tau^{2}), && \forall \nsamp,  \forall \tau \geq 0\\
\notag	\Lipdev(\radius\ \bar{g}(\radius^{2}/2)) &= 0, && \forall \nsamp.  
\end{align}
In other words, {\bf C2} holds with $c=\radius^{2}/\sqrt{8}$ and $T=+\infty$, and {\bf C1} holds with $L_{\PP}(\bar{g}) \leq \radius \bar{g}(\radius^{2}/2))$.
\end{lemma}
\begin{IEEEproof}
For any $\Db$, the random variables $y_{i} = f_{\signal_{i}}(\Db)$ satisfy $0 \leq y_{i} \leq \tfrac{1}{2} \|\signal\|_{2}^{2} \leq \tfrac{\radius^{2}}{2}$ almost surely. Applying Hoeffding's inequality yields~\eqref{eq:ConcentrationSphere}.
We conclude by observing that since $\|\signal_{i}\|_{2} \leq \radius$ almost surely, we have $L_{\Signal}(\bar{g}) \leq \radius\ \bar{g}(\radius^{2}/2)$ almost surely.
\end{IEEEproof}

\begin{lemma}\label{le:ConcentrationSubGaussian}
Assume there is a constant $A>0$ such that $\PP \in \Pfrak_{A}$ and that the penalty function $g$ satisfies {\bf A1} (non-negativity) and {\bf A4} ($g(\mathbf{0}) = 0$). Then
\begin{align*}
\Expdev(12A\tau)  &\leq 2 \exp(-\nsamp\tau^{2}),\ \forall 0 \leq \tau \leq 1.
\end{align*}
In other words, {\bf C2} holds with $c=12A$ and $T=1$.

Moreover, in all examples of penalties $g$ considered in Table~\ref{tab:EnveloppeBarG}, the growth of $t \bar{g}(t^{2}/2)$ for large $t$ is at most polynomial, so that $L_{\PP}(\bar{g}) < +\infty$, and {\bf C1} holds.
\end{lemma}
\begin{IEEEproof}
We follow the argument in \cite{jenatton:hal-00737152}. This primarily results from Bernstein's inequality, see, e.g., \cite{Bennett:1962vd}.
\begin{lemma}[Bernstein's Inequality]\label{lem:bernstein}
Let $\{z_j\}_{j\in\SET{\nsamp}}$ be independent, zero-mean random variables. 
If there exist $v,M >0$ such that for any integer $q \geq 2$ and any $j\in\SET{n}$,
it holds
\begin{equation*}
\EE[ |z_j|^q ] \leq \frac{q!}{2} M^{q-2} v^2, 
\end{equation*}
then we have for any $\gamma \geq 0$,
\begin{equation*}
\PP\Big( \tfrac{1}{\nsamp} \sum_{j=1}^\nsamp z_j > \gamma \Big)\leq 
\exp\Big( -\frac{\nsamp \gamma^2}{2  (v^2 + M \gamma)} \Big).
\end{equation*}
In particular, for any $\tau \leq \frac{v}{2M}$, we have
\begin{equation*}
\PP\Big( \tfrac{1}{n}\sum_{j=1}^n z_j > 2v\tau \Big)\leq 
\exp\big( -\nsamp\tau^2 \big).
\end{equation*}
\end{lemma}
We can now state a simplification of \cite[Lemma 24]{jenatton:hal-00737152}.

\begin{lemma}
\label{lem:truncated_moments}
Let $y$ be a random variable and assume there is a constant $B>0$ such that, for any $t \geq 1$,
\begin{equation}
\label{eq:DefExponentialDecay}
\PP\left( |y| > B t \right) \leq \exp(-t).
\end{equation}
Then, for any $u \geq 1$, any integer $q \geq 1$, and $0<p\leq 1$, we have
\begin{align}
\label{eq:qExpectation}
\EE\Big[ |y|^{pq} \Big]
&\leq q! \Big[ B^p u \Big]^q  
[1+\exp(3-u)]\\
\label{eq:qMomentDeviation}
\EE\Big[ \Big| |y|^p  - \EE\big[ |y|^p\big]  \Big|^q  \Big]
&\leq q! \Big[ 2B^p u \Big]^q  
[1+\exp(3-u)].
\end{align}
\end{lemma}
To keep the flow of the paper we postpone the proof to the appendix. We now have the tools to state a simplified version of \cite[Corollary 6]{jenatton:hal-00737152} which suits our needs.
\begin{corollary}\label{cor:concentration_subgaussian}
Consider $\nsamp$ independent draws $\{ y_{i} \}_{ i\in\SET{\nsamp} }$ satisfying the hypothesis~\eqref{eq:DefExponentialDecay} for some $B>0$.
Then, for any $0<p\leq1$ and $0\leq \tau \leq 1$,  we have
\begin{align}
\label{eq:concentration_subgaussian_expectation}
\EE\{|y_{i}|^p\}
\leq 
6 B^{p} &\\
\label{eq:concentration_subgaussian_deviation}
\PP\left(\Big|\tfrac{1}{\nsamp} \sum_{i=1}^{\nsamp} \left(|y_{i}|^{p} -\EE\ |y_{i}|^p\right)\Big|
\geq 
24 B^{p} \cdot \tau
\right) 
& \leq  2 \exp(-\nsamp\tau^{2})
\end{align}
\end{corollary}
\begin{IEEEproof}
We apply Lemma~\ref{lem:truncated_moments} with $u=3$. For $q=1$, Equation~\eqref{eq:qExpectation} yields~\eqref{eq:concentration_subgaussian_expectation}.
For $q \geq 2$, Equation~\eqref{lem:truncated_moments} shows that we can apply Bernstein's inequality (Lemma~\ref{lem:bernstein}) with $z_{i} = |y_{i}|^{p} -\EE\ |y_{i}|^p$, $M = 2B^p u = 6 B^{p}$ and $v = \sqrt{2}M\sqrt{1+e^{3-u}} = 2M= 12 B^p$. This shows that for $0 \leq \tau \leq v/2M=1$ we have~\eqref{eq:concentration_subgaussian_deviation}.
\end{IEEEproof}

We are now ready to prove Lemma~\ref{le:ConcentrationSubGaussian}. First, 
we notice that for any $\Db$, $0 \leq y_{i} = f_{\signal}(\Db) \leq \tfrac{1}{2} \|\signal\|_{2}^{2}$, hence we have
\begin{equation*}
\sup_{\Db \in \Cfrak} \PP\left(f_{\signal}(\Db) \geq \tfrac{A}{2} t\right) \leq \exp(-t),\ \forall t \geq 1.
\end{equation*}
Applying Corollary~\ref{cor:concentration_subgaussian} with $B=A/2$ and $p=1$ yields $\Expdev( 12 A \tau) \leq 2 \exp(-\nsamp \gamma^{2})$ for $0 \leq \tau \leq 1$. 
\end{IEEEproof}

\section{Constraint set structures and covering numbers}
\label{sec:Covering}

Dictionaries learned with standard methods are unstructured matrices $\Db \in \R{\sdim \times \cdim}$ that allow factored representations of the signals of interest. However, in practice the dimension of the signals which are being represented and consequently the possible dictionaries' dimensions are inherently restricted by limited memory and limited computational resources. Furthermore, when used within signal reconstruction algorithms where many matrix vector multiplications have to be performed, those dictionaries are computationally expensive to apply. In the following we first describe various classes of dictionaries and their motivation in signal processing. Then we turn to the estimation of their covering numbers.

\subsection{Unstructured Dictionaries with Unit Norm Atoms} Imposing unit norm on the atoms is a common technique to avoid trivial solutions for the learning problem. An exemplary learning algorithm that employs this kind of restriction is the famous K-SVD algorithm proposed in \cite{dl:aharon:2006}. All following dictionary structures base upon this restriction. The class of dictionaries of this form is a product of spheres defined as
\begin{equation*}
	\Cfrak(\sdim,\cdim) \defin \{ \Db \in \R{\sdim \times \cdim} \,:\, \operatorname{ddiag}(\Db^\top \Db) = \mathbf{I}_\cdim \}.
\end{equation*}
The operator $\operatorname{ddiag}$ maps matrices $\R{\cdim \times \cdim}$ to $\R{\cdim \times \cdim}$ by leaving the diagonal entries unaltered while setting all others to zero.

In NMF, this dictionary structure is further limited by requiring non-negative entries, yielding the class $\CNMF \cap \Cfrak(\sdim,\cdim)$, or $\Cfrak_{\textrm{NMF}\& \ell^{1}}$ if the $\ell^{1}$ normalization is preferred.

\begin{lemma}[Covering number bound for $\Cfrak(\sdim,\cdim)$]\label{lem:cover}
For the Euclidean metric, $\Sphere^{\sdim-1}$ the unit sphere in $\R{\sdim}$ and for any $\epsilon >0$, we have 
\begin{equation*}
	\Ncal(\Sphere^{\sdim-1},\epsilon) \leq \big(1+2/\epsilon\big)^\sdim.
\end{equation*}
Moreover, for the metric induced on $\R{\sdim\times \cdim}$ by $\|\cdot\|_{1 \to 2}$, and for any $\epsilon > 0$, we have 
\begin{equation*}
	\Ncal(\Cfrak(\sdim,\cdim),\epsilon) \leq \big(1+2/\epsilon\big)^{\sdim\cdim}.
\end{equation*}
\end{lemma}
\begin{IEEEproof}
Lemma 2 in \cite{Vershynin2010} gives the first conclusion for the sphere in $\R{\sdim}$.
As for the second result,
remember that the set $\Cfrak(\sdim,\cdim)$ is a Cartesian product of $\cdim$ $(\sdim\!-\!1)$-spheres.
\end{IEEEproof}

\subsection{Sparse Dictionaries} The authors of \cite{dl:rubinstein:2010} propose a technique to learn so-called sparse dictionaries. A sparse dictionary is the representation of a learned dictionary $\Db$ and a given base dictionary $\mathbf{\Theta}$ (e.g., DCT, ODCT, Wavelet) with the sparse matrix $\bar{\Db}$ via $\Db = \mathbf{\Theta} \bar{\Db}$.
This learning algorithm offers a reduction in training complexity while aiming at maintaining the data modeling abilities of unconstrained dictionary learning.
We show in the appendix that:
\begin{equation*}
	\Ncal(\Cfrak_{\mathrm{sparse}}(\sdim,\cdim,s,\mathbf{\Theta}),\epsilon) \leq \left( \tbinom{\sdim}{s} \left(1+2/\epsilon\right)^s \right)^\cdim.
\end{equation*}	

\subsection{Orthogonal Dictionaries} Orthogonal dictionaries are square matrices belonging to
\begin{equation}
\label{eq:def_Dorth}
	\Orth(\cdim) \defin \{ \Db \in \R{\cdim\times\cdim} \,:\, \Db^\top\Db = \mathbf{I}_\cdim, \det(\Db) = 1\}.
\end{equation}
They represent the simplest form of dictionaries that allow a unique representation of the signal. Computing the inner product of the signal and the atoms of the dictionary provides the representation coefficients. We show in the appendix that:
\begin{equation*}
	\Ncal(\Cfrak_\mathrm{orth}(\cdim),\epsilon) \leq \left( 3\pi e^\pi/\epsilon\right)^{\cdim(\cdim-1)/2}.
\end{equation*}
\subsection{Stiefel manifold} In PCA (see Section~\ref{sec:WorkedExamples}), the dictionary is a matrix consisting of $\cdim$ orthonormal columns where $\cdim \leq \sdim$, i.e., $\Db$ is a tall matrix. This constraint set is the so called Stiefel manifold
\begin{equation}
\label{eq:def_stiefel}
	\ST(\sdim,\cdim) \defin \{ \Db \in \R{\sdim \times \cdim} \,:\, \Db^\top\Db = \mathbf{I}_\cdim \}.
\end{equation}
We show in the appendix that:
\begin{equation*}
	\Ncal(\Cfrak_{\mathrm{St}}(\cdim,\sdim),\epsilon) \leq \left( 3\pi e^\pi/\epsilon \right)^{\sdim \cdim - \cdim(\cdim+1)/2}.
\end{equation*}
\subsection{Separable Dictionaries} In \cite{cvpr:hawe:2013} a dictionary learning scheme is developed that learns a separable dictionary for sparse image representations:  instead of  sparsely representing $\signal = \operatorname{vec}(\mathbf{Y})$ as $\signal = \Db \coeff$, where $\mathbf{Y}$ is a matrix containing an image, one directly represents $\mathbf{Y}$ sparsely as $\mathbf{B}$ via two dictionaries $\Db_1$, $\Db_2$ by the equation 
\begin{equation}\label{eq:sep_dic_l}
	\mathbf{Y} = \Db_1^{} \mathbf{B} \Db_2^\top.
\end{equation}
Note that in order to relate this method to standard dictionary learning techniques Equation~\eqref{eq:sep_dic_l} can be rewritten as 
\begin{equation*}
	\operatorname{vec}(\mathbf{Y}) = \left(\Db_2 \otimes \Db_1\right) \cdot \operatorname{vec}(\mathbf{B}).
\end{equation*}
We show in the appendix that:
\begin{equation*}
	\Ncal(\Cfrak_{\mathrm{sep}},\epsilon) \leq \left( 3/\epsilon \right)^{\sum_i\sdim_i \cdim_i}.
\end{equation*}

\subsection{Other Tensor Dictionaries}
Similar to the structure of Separable dictionaries we can define other constraint sets that are combinations of dictionary structures. Examples for this are tensor products of Stiefel matrices defined as
\begin{equation}
\label{eq:def_stiefel_tensor}
	\Cfrak_{\otimes \mathrm{St}} \defin \Cfrak_\mathrm{St}(\sdim_1,\cdim_1)\otimes \ldots \otimes \Cfrak_\mathrm{St}(\sdim_z,\cdim_z)
\end{equation}
which are used for Higher-order SVD. For the sake of readability the dimensions are omitted in this definition.

We show in the appendix that:
\begin{equation*}
	\Ncal(\Cfrak_{\otimes \mathrm{St}},\epsilon) \leq \left( 3\pi e^\pi/\epsilon \right)^{\sum_i \sdim_i \cdim_i - \cdim_i (\cdim_i +1)/2}.
\end{equation*}

\section{Worked Examples}
\label{sec:WorkedExamples}

In this section we propose some settings that fit to our framework and provide the corresponding sample complexity estimates, i.e., the value of $\beta$. 
\begin{example}[PCA]
PCA can be expressed as the optimization problem
\begin{equation*}
\label{eq:PCA_cost}
	\min_{\Ab;\Db \in \Cfrak_\mathrm{St}(\sdim,\cdim)}\tfrac{1}{2} \|\mathbf{X} - \Db\Ab\|_F^2 + \tfrac{1}{\nsamp}\sum g(\coeff_i)
\end{equation*}
with the penalty function
%
$g = \chi_{\cdim\textrm{-sparse}}$.
%
Since $\coeff_i \in \R{\cdim}$, the penalty function is equivalent to $0$. Its sole purpose is to shape the problem in a way that fits into the general framework~\eqref{eq:basic_dict}. The constraint set is $\Cfrak_\mathrm{St}(\sdim,\cdim)$, the set of $\sdim\!\times\!\cdim$-matrices with orthonormal columns. 

\paragraph{Training data in the unit ball}
Initially, we consider a distribution $\PP$ in the unit ball ($\radius=1$, therefore, $c=1/\sqrt{8}$, cf. Lemma~\ref{le:ConcentrationSphere}). 
Since $\Cfrak_\mathrm{St}(\sdim,\cdim) \subset \Ck$ with $k=\cdim$ and $\delta=0$, the Lipschitz constant is $L=2\sqrt{\cdim}$, cf.~Table~\ref{tab:EnveloppeBarG}. Table~\ref{tab:Covering} yields $h=\sdim\cdim-\cdim(\cdim+1)/2$ and $C=3 \pi e^\pi$. Hence, the constant driving the sample complexity is
\begin{equation*}
	\beta_{\mathrm{PCA}} = \left(\sdim\cdim-\tfrac{\cdim(\cdim+1)}{2}\right)\cdot \log(12\pi e^\pi \sqrt{8\cdim}).
\end{equation*}
Compared to existing sample complexity results for PCA \cite{ShaweTaylor:2005eu,blanchard:2007}, this seems rather pessimistic: in \cite{blanchard:2007} the sample complexity is driven by $\beta \propto \cdim$, and we lose a large dimension-dependent $\sdim$ factor\footnote{See Section~\ref{sec:sample_alt} for  techniques to handle this high-dimensional scaling.} as well as an additional mild $\log \nsamp$ factor. 
In fact, the assumption that the data is distributed in the unit ball is crucial in the results of \cite{ShaweTaylor:2005eu,blanchard:2007} as it allows the use of the McDiarmid concentration inequality,
 while, in our general context, we exploit a standard but somewhat crude argument using union bounds and covering numbers. This however means we can handle data distributions not restricted to live on the unit ball, as we will see next.

\end{example}

\paragraph{Sub-Gaussian training data}
The sub-optimality of our results for PCA with respect to state of the art is probably the price to pay for the generality of the approach. 
For example, in contrast to previous work, our approach provides results for other distributions. For example, given the distribution $\PP \in \Pfrak_{A}$, we obtain the constant $c=12A$ (Lemma~\ref{le:ConcentrationSubGaussian}), while all other constants remain the same. Thus, we get
\begin{equation}
	\beta_{\mathrm{PCA}} = \left(\sdim\cdim-\tfrac{\cdim(\cdim+1)}{2}\right)\cdot \max\left(\log\left(\tfrac{\pi e^\pi \sqrt{\cdim}}{A}\right),1\right),
\end{equation}

\begin{example}[Sparse coding, training data in the unit ball]
Consider $\PP$ a distribution in the unit ball ($\radius=1$).  We have $c=1/\sqrt{8}$ (Lemma~\ref{le:ConcentrationSphere}), and Table~\ref{tab:EnveloppeBarG} immediately provides values of the Lipschitz constant $L=\bar{g}(1/2)$ such that $\Lipdev(L)=0$. For example, for $0<p<\infty$
\begin{itemize}
	\item for $g_{\chi_{p}}(\coeff) = \chi_{\|\coeff\|_{p} \leq \lambda}$, we have $L_{p}= d^{(1-1/p)_{+}}\lambda$;
	\item for $g_{\chi_{0}}(\coeff) = \chi_{\|\coeff\|_{0} \leq k}$, we have $L_{0} = 2\sqrt{\tfrac{k}{1-\delta}}$ 
\end{itemize}

For $\Cfrak = \Cfrak(\sdim,\cdim)$ (resp. $\Cfrak = \Ck$), Table~\ref{tab:Covering} provides $\cvdim=\sdim\cdim$ and $C=3$, and
Theorem~\ref{th:MainTheorem} yields 
\begin{align*}
	\beta_{\chi_{p}} &= \sdim\cdim \cdot \max\left(\log 6 \sqrt{8} \lambda d^{(1-1/p)_{+}},1\right),\\
	\beta_{\chi_{0}} &= \sdim \cdim \cdot \log 12 \sqrt{8k/(1-\delta)}.
\end{align*}

The cases $g_{\chi_{1}}$ and $g_{\chi_{0}}$ are dealt with by Vainsencher {\em et al.} \cite[Theorem~7]{vainsencher:2010} who provide a comparable result,
however restricted to distributions on the unit sphere whereas our framework allows for more complex distributions. Moreover while they make assumptions on the cumulative coherence of $\Db$ for their results with the $\lzero$ constraint, we only rely on the restricted isometry property. 

The bound of Maurer and Pontil \cite{MaurerPontil} for $g_{\chi_{p}}$, $p\geq1$, $\lambda = 1$, is of the order $\beta'_{p} \propto \cdim^{4-2/p}$. We extend it to $p<1$ as well, and improve the case $p=1$ as soon as $\sdim \leq \cdim$.

In addition to the above penalty functions associated to the indicator functions of the $\ell^{1}$ (resp. $\lzero$) ball, our results allow dealing with the $\ell^{1}$ norm itself --a very popular proxy for the $\ell^{0}$ ``norm''-- or its powers as penalty functions. For $\Cfrak = \Cfrak(\sdim,\cdim)$ (resp. $\Cfrak = \Ck$), the sparsity measures
\begin{itemize}
	\item $g_{\ell^{1}}(\coeff) = \|\coeff\|_{1}/\lambda$, yielding $L=\tfrac{\lambda}{2}$;
	\item $g_{(\ell^{1})^2}(\coeff) = \|\coeff\|_{1}^{2}/\lambda^{2}$, yielding $L = \tfrac{\lambda}{\sqrt{2}}$;
\end{itemize}
are associated to the sample complexities
\begin{align*}
	\beta_{\ell^{1}} &= \sdim\cdim \cdot \max\left(\log3\sqrt{8}\lambda,1\right)\\
	\beta_{(\ell^{1})^2} &= \sdim\cdim \cdot \max\left(\log 12\lambda,1\right)
\end{align*}
for probability distributions in the unit ball.
\end{example}

\begin{example}[Sparse coding, sub-Gaussian training data]
For a distribution $\PP$ satisfying assumption~\eqref{eq:ConcentrationSubGaussian2} we have $c=12A$. 
For $\Cfrak=\Cfrak(\sdim,\cdim)$ we again have the constants $C=3$, $h=\sdim\cdim$ and Theorem~\ref{th:MainTheorem} then yields sample complexity for $g_{(\ell^{1})^{2}}$
\begin{equation*}
	\beta = \sdim \cdim \cdot \max\left(\log\tfrac{\lambda}{2\sqrt{2}A},1\right).
\end{equation*}
\end{example}

\begin{example}[``Doubly sparse'' coding, training data in the unit ball]
Our results also allow to deal with other  classes of structured dictionaries, such as the sparse dictionary $\Cfrak_{\mathrm{sparse}}(\sdim,\cdim)$ defined in Equation~\eqref{eq:def_sparse_dic}, yielding $\cvdim = s\cdim$ and $C = 3 \tbinom{\sdim}{s}^{1/\cdim}$. With $g_{\chi_{1}}$ (respectively $g_{\chi_{0}}$ with a restriction to $\Cfrak_{\mathrm{sparse}}(\sdim,\cdim) \cap \Ck$), using Stirling's approximation, Theorem~\ref{th:MainTheorem} yields
\begin{align*}
	\beta_{\chi_{1}}^{s} & \leq s \cdim \cdot \max\left(\log 6 \sqrt{8}\lambda + \log \tfrac{\sdim e}{s},1\right)\\
	\beta_{\chi_{0}}^{s} & \leq s \cdim \cdot \left(\log 12 \sqrt{8k/(1-\delta)} + \log \tfrac{\sdim e}{s}\right).
\end{align*}

\end{example}

\begin{example}[Non-negative Matrix Factorization, training data in the unit ball]
In Non-negative Matrix Factorization \emph{(NMF)} (cf. \cite{seung:2001algorithms}) a matrix $\Signal \in \R{\sdim \times \nsamp}$ is decomposed into a product of two matrices with non-negative entries. It can be expressed as the optimization problem
\begin{equation}
	\min_{\Coeff;\Db\in\Cfrak_{\textrm{NMF} \& \ell^{1}}} \tfrac{1}{2} \|\Signal - \Db \Coeff\|_F^2 + \sum_{i=1}^{\nsamp} \chiNMF(\coeff_i).
\end{equation}
Since the constraint set $\Cfrak_{\textrm{NMF} \& \ell^{1}}$ is a subset of $\CNMF(1/\sdim\cdim)$,
the penalty $g=\chiNMF$ and the set $\Cfrak_{\textrm{NMF} \& \ell^{1}}$ satisfy {\bf B1-B3} with $\kappa=1/\sdim\cdim$, and Table~\ref{tab:EnveloppeBarG}
gives the Lipschitz constant $L=2\sqrt{\sdim\cdim}$. 
Moreover, since $\Cfrak_{\textrm{NMF} \& \ell^{1}} \subset \Cfrak_{\textrm{ball}}(\sdim,\cdim)$,  where
\begin{equation*}
	\Cfrak_\mathrm{ball}(\sdim,\cdim) \defin \{\Db \in \R{\sdim\times \cdim} \,:\, \|\db_i\|_2 \leq 1\}
\end{equation*}
has the same covering number as $\Cfrak(\sdim,\cdim)$, we get $\Ncal(\Cfrak,\epsilon) \leq \left(\frac{3}{\epsilon}\right)^{\sdim\cdim}$  as a direct consequence of the covering number bound for the product of spheres. This yields the constants $C=3$ and $h=\sdim\cdim$, and together with the constant $c=1/\sqrt{8}$, which we get for distributions on the unit sphere, we obtain
\begin{equation}
	\beta_{\mathrm{NMF}} = \sdim \cdim \cdot \log 12 \sqrt{8\sdim\cdim}.
\end{equation}
In a similar setting Maurer and Pontil \cite{MaurerPontil} obtain a sample complexity estimate $\beta \propto \cdim^{3}$ which has the advantage of being independent of the signal dimension $\sdim$. Yet, our result seems sharper unless the signal dimension $\sdim$ exceeds  $\cdim^{2}$, in which case the approach sketched in Section~\ref{sec:sample_alt} could be followed to recover improved estimates in this high-dimensional scaling.
%
\end{example}
\begin{example}[$K$-means clustering, training data in the unit ball]
Given a set of $\nsamp$ data points $K$-means clustering algorithms learn a dictionary with $K$ columns and assign each data point to one of these columns while minimizing the sum of squared distances. These atoms represent the $K$ distinct classes.
In our notation so far $\cdim=K$. Then the problem of $K$-means clustering can be expressed as the minimization problem
\begin{equation}
	\min_{\Coeff;\Db \in \R{\sdim \times K}}  \tfrac{1}{2} \|\Signal - \Db\Coeff\|_F^2 + \sum_{i=1}^{n} g(\coeff_i)
\end{equation}
with the penalty function
\begin{equation}
	g(\coeff) = \chi_{\|\coeff\|_0=1} + \chi_{\sum_k\alpha_k=1}.
\end{equation}
This penalty function ensures that all columns of $\Coeff$ have exactly one entry with the value $1$. 
The centers of the clusters are represented in the columns of $\Db$. 

This penalty function fulfills \textbf{A1-A3}, and its auxiliary function is $\bar{g}(t) \equiv 1$ for all $t \in \R{}$. However, $g$ does not fulfill \textbf{A4}. This has the consequence that we cannot merely apply Corollary~\ref{le:ConstantsWithNormPenalties}. Instead, we have to rely on the preceding Lemma~\ref{le:ConstantsWithNormPenalties_A1A3} which leaves us with the upper bound $\tfrac{1}{\nsamp} \sum_{i=1}^\nsamp \sqrt{2 f_{\signal_i}(\Db)}\cdot \bar{g}(f_{\signal_i}(\Db)) = \tfrac{1}{\nsamp} \sum_{i=1}^\nsamp \sqrt{2 f_{\signal_i}(\Db)}$ for the Lipschitz constant. To proceed, we recall the definition $f_{\signal}(\Db) = \inf_{\coeff} \tfrac{1}{2}\|\signal - \Db\coeff\|_2^2 + g(\coeff)$. For the function $\tfrac{1}{2}\|\signal - \Db\coeff\|_2^2 + g(\coeff)$ to be finite, the coefficient vector $\coeff$ has to fulfill $g(\coeff)=0$. Due to the construction of $g$, there is only a finite number of possible choices for $\coeff$ (namely $\coeff$ is a standard basis vector in $\R{d}$), and the Lipschitz constant can be upper bounded by
%
\begin{equation*}
	L_\Signal(\Db) \leq \tfrac{1}{\nsamp}\sum_{i=1}^{\nsamp}\max_{j=1,\ldots,K} \| \signal_i - \db_j \|_2 
	\leq \max_{i,j}\| \signal_i - \db_j \|_2.
\end{equation*}

Restricting our analysis to the case of data samples $\signal_{i}$ lying within the unit ball ($\|\signal_{i}\|_{2} \leq 1$), the optimum dictionary coincides with the optimum where the matrix $\Db$ is constrained to belong to 
%
	$\Cfrak_\mathrm{ball}(\sdim,K)$.
It follows that the Lipschitz constant is simply bounded by $L=2$. Moreover, $\Cfrak_\mathrm{ball}(\sdim,K)$ has the same covering number as $\Cfrak(\sdim,K)$.
From the structural constraints to the dictionary and the signal distribution we obtain the constants $L=2$, $C=3$, $h=\sdim K$, $c=1/\sqrt{8}$. 
The above implies the constant
\begin{equation*}
	\beta_{K\mathrm{-means}} = \sdim K \cdot \log (12 \sqrt{8}).
\end{equation*}

The literature on the sample complexity of $K$-means is abundant. Following the early work of Pollard \cite{Pollard:1982km}, a series of authors have established worst-case lower bounds on the excess risk decaying at best as $1/\sqrt{\nsamp}$ \cite{Linder:2001jl,bartlett1998minimax,Antos:2005ik}, as well as upper bounds of the order of $O(1/\sqrt{\nsamp})$ (for a given dimension $\sdim$) \cite{Linder:1994il}. More concrete estimates, where the role of $K$ and $\sdim$ is explicated, are needed to compare with our sample complexity estimate. Bartlett {\em et al.}  \cite{bartlett1998minimax} bound the excess risk by a constant times $\min\left(\sqrt{K\sdim/\nsamp},\sqrt{K^{1-2/\sdim} \sdim \log \nsamp/\nsamp}\right)$ which, for large $\sdim$ and arbitrary $K$, is essentially matched by our estimate. Estimates independent of the dimension $\sdim$ have been obtained by Biau {\em et al.} \cite{biau2008performance} (see also \cite{MaurerPontil}) of the order $K/\sqrt{\nsamp}$. These correspond to $\beta \propto K^{2}$, hence our result seems sharper when $\sdim \lesssim K$, i.e., when the number of clusters exceeds the dimension.  

\end{example}
\begin{example}[Higher-order SVD]
A natural extension of PCA is the so called Higher-order SVD \emph{(HOSVD)} or multilinear SVD introduced in \cite{de2000multilinear}.
Let the training signals $\signal_{i}$ be real $z$-tensors, i.e., $\Signal$ is an element of $\R{\sdim_1 \times \sdim_2 \times \ldots \times \sdim_z \times \nsamp}$ with $\Signal = \left[ \Signal_1\ \ldots\ \Signal_\nsamp \right]$ where $\Signal_i \in \R{\sdim_1 \times \sdim_2 \times \ldots \times \sdim_z}$. 
The HOSVD $\Signal$ is obtained by solving the minimization problem
\begin{equation*}
	\min_{\Ab, \Db \in \Cfrak_{\otimes\mathrm{St}}} \tfrac{1}{2} \sum_{i=1}^\nsamp \|\Signal_i - \Db \circ \Ab_i\|_F^2.
\end{equation*}
In this equation $\Db$ is an element of the constraint set $\Cfrak_{\otimes\mathrm{St}}$ as defined in \eqref{eq:def_stiefel_tensor} and therefore a concatenation of Stiefel matrices $\Db_1,\ldots,\Db_z$ with $\Db_i \in \ST(\sdim_i,\cdim_i)$, and the coefficient tensor $\Ab \in \R{\cdim_1 \times \ldots \cdim_z \times \nsamp}$ has the form $\Ab = \left[\Ab_1\ \ldots\ \Ab_\nsamp \right]$ with $\Ab_i \in \R{\cdim_1 \times \ldots \cdim_z}$. The operator $\circ$ denotes the operation
\begin{equation*}
	\Db \circ \Ab_i \defin \Ab_i \times_1 \Db_1 \times_2 \Db_2 \ldots \times_z \Db_z,
\end{equation*}
where $\times_N$ denotes the so-called $N$-mode product introduced in \cite{de2000multilinear}.
Just as for PCA the trivial penalty $G(\Ab)\equiv 0$ can be written using $g(\coeff) = \chi_{\|\coeff\|_{0} \leq \prod_{i=1}^{z}d_{i}}$ where $\coeff_{i}$ is the vector resulting from the unfolding of the $i$-th component of $\Ab$, $1 \leq i \leq \nsamp$, along its rightmost dimension.

Table~\ref{tab:Covering} yields the constants 
$C=3 \pi e^\pi$ and $h=\sum_{i=1}^{z}\sdim_i\cdim_i - \cdim_i(\cdim_i+1)/2$. To obtain the Lipschitz constant $L$ we unfold the tensor $\Ab$ to the matrix $\R{(\prod_{i=1}^z \cdim_i) \times \nsamp}$ and then use a penalty definition similar to the one used for PCA. This yields the constant $L=2\cdot\sqrt{\prod_{i=1}^z \cdim_i}$. For distributions $\PP$ in the ball with $\radius=1$, Theorem~\ref{th:MainTheorem} provides 
\begin{equation*}\begin{split}
	\beta_{\mathrm{HOSVD}} = \left(\sum_{i=1}^{z}\sdim_i\cdim_i - \tfrac{\cdim_i(\cdim_i+1)}{2}\right) \cdot \log\left(12\pi e^\pi \sqrt{8\prod\limits_{i=1}^z \cdim_i}\right).
\end{split}\end{equation*}
In light of state of the art results \cite{blanchard:2007} on PCA, this apparently new sample complexity estimate is probably quite pessimistic. 
\end{example}

\GOLrev{
\section{High-dimensional settings}
\label{sec:sample_alt}
When $\sdim$ gets larger, the reasoning in Section~\ref{sec:Concentration} based on covering numbers of the set of admissible dictionaries $\Cfrak$ leads to unfavorable scaling with respect to $\sdim$.
This factor is introduced through the $\epsilon$-net argument following Lemma~\ref{le:CoveringArgumentGeneric}.
In order to avoid the issues that arise in the high-dimensional setting, we could envision using an alternative to Lemma~\ref{le:OneSidedLipschitz} which served as a precursor to the discussion of the Lipschitz property. Such an alternative is proposed in Lemma~\ref{le:OneSidedLipschitz_alt}, where we derive a Lipschitz property of $F_\Signal$ w.r.t.\ the Frobenius norms of $\Db^\top\Db$ and $\Db^\top\Signal$ which are in $\R{\cdim \times \cdim}$ and $\R{\cdim \times \nsamp}$, respectively. Covering net arguments revolving around these quantities clearly become independent of the signal dimension $\sdim$, but would yield difficulties due to their dependence on the draw of $\Signal$. Providing concrete results for this cases is beyond the scope of this paper, instead, we sketch below an alternative approach.\\
Another option to handle the high-dimensional setting is to extend the approach proposed in \cite{MaurerPontil}. We provide a rough outline below. First, recall that the initial problem is to provide an upper bound for the expression
\begin{equation*}
	\sup_{\Db \in \Cfrak} \left( \tfrac{1}{\nsamp} \sum_{i=1}^{\nsamp} f_{\signal_i}(\Db) - \mathbb{E}_\signal f_\signal(\Db)\right).
\end{equation*}
We can use McDiarmid's inequality in combination with a symmetrization argument and Rademacher averages to provide the upper bound
%
\begin{align*}
	\sup_{\Db \in \Cfrak} \left( \tfrac{1}{\nsamp} \sum_{i=1}^\nsamp f_{\signal_i}(\Db) - \mathbb{E}_{\signal} f_{\signal}(\Db) \right) 
	= &\text{C}_1\cdot \left[\mathbb{E}_{\eta} \sup_{\Db \in \Cfrak} A_\Db(\eta)\right] \\
	&+ \text{C}_2 \cdot \sqrt{2t/\nsamp}
\end{align*}
with probability at least $1-e^{-t}$ and the Gaussian process $A_\Db(\eta) = \tfrac{1}{n}\sum_{i=1}^\nsamp \eta_i f_{\signal_i}(\Db)$ where the variables $\eta_i$ are i.i.d.\ standard Gaussian. 
Therefore, it remains to find an upper bound for $\mathbb{E}_{\eta} \sup_{\Db \in \Cfrak} A_\Db(\eta)$. Slepian's Lemma states that for any Gaussian process $B_\Db$ that fulfills the condition $\mathbb{E}_\eta |A_\Db(\eta) - A_{\Db'}(\eta)|^2 \leq \mathbb{E}_{\xi}|B_\Db(\xi) - B_{\Db'}(\xi)|^2$ the bound $\mathbb{E}_{\eta} [\sup_{\Db \in \Cfrak} A_\Db(\eta)] \leq \mathbb{E}_{\xi} [\sup_{\Db \in \Cfrak} B_\Db(\xi)]$ holds. In the case at hand we can observe that $\mathbb{E}_\eta |A_\Db(\eta) - A_{\Db'}(\eta)|^2 = \tfrac{1}{\nsamp^2} \sum_{i=1}^{\nsamp} |f_{\signal_i}(\Db) - f_{\signal_i}(\Db')|^2$. Lemma~\ref{le:OneSidedLipschitz_alt} yields the upper bound
\begin{align}
\label{eq:GaussP2}
	|f_{\signal}(\Db) - f_{\signal}(\Db')| \leq & C^{(1)}_{\Signal}\|\Db^\top \Db - (\Db')^\top \Db'\|_F\nonumber\\
&	 + C^{(2)}_{\Signal}\|(\Db - \Db')^\top \signal\|_2,
\end{align}
which implies that the Gaussian process
\begin{equation}
	\label{eq:GaussP3}
	B_{\Db}(\xi) = C'\langle \Db^\top \Db, \xi^{(1)} \rangle_F
									+ C''\langle \Db^\top \signal, \xi^{(2)} \rangle_2
\end{equation}
fulfills the condition of Slepian's Lemma for appropriate choices of the scalars $C',C''$.
Here, the first scalar product operates in $\R{\cdim \times \cdim}$, whereas the second one is defined on $\R{\cdim}$. Thus, computing the supremum over all $\Db$ would result in an expression which is independent of the signal dimension $\sdim$. Especially in the case of PCA, the final sample complexity result would benefit since the first term in Equation~\eqref{eq:GaussP2}, and thereby the first term in \eqref{eq:GaussP3}, vanishes, yielding a tighter bound. 
However, it is not within the scope of this work to provide concrete results for this alternative approach. 
}

\section{Conclusion and discussion}

We proposed a general framework to determine the sample complexity of dictionary learning and related matrix factorization problems. The generality of the framework makes it applicable for a variety of structure constraints, penalty functions, and signal distributions beyond previous work. In particular, it covers formulations such as principal component analysis, sparse dictionary learning, non-negative matrix factorization, or $K$-means clustering, for which we provide sample complexity bounds in the worked examples section. 

To keep the exposition of our results simple, we focused on matrix factorization where the data fidelity term is expressed with squared $\ell^{2}$ norm. A straightforward adaptation of the computation of the Lipschitz constant of Section~\ref{sec:Lipschitz} can easily be made to achieve similar results with the plain $\ell^{2}$ norm \cite{vainsencher:2010}, and easy adaptations can be envisioned when the data fidelity term is $\ell(\signal-\Db\coeff)$ where the gradient $\nabla \ell(\cdot)$ is Lipschitz. More general loss functions are briefly discussed in Section~\ref{sec:other losses}.

The obtained sample complexity results applied to sparse coding extend those of Maurer and Pontil \cite{MaurerPontil} and Vainsencher {\em et al.} \cite{vainsencher:2010} in primarily two ways. 

First, we relax the assumption that the training data lives in the unit ball \cite{MaurerPontil} or even the unit Euclidean sphere \cite{vainsencher:2010} by showing that it is sufficient to have sufficient decay of the probability of drawing training samples with ``large'' norm. This is essentially achieved by replacing Hoeffding's inequality with a more refined Bernstein inequality argument.

Second, and more importantly, we handle penalty functions $g$ beyond indicator functions of compact sets \cite[Theorem 1]{MaurerPontil}, or of sets $\mathcal{K}$ such that $\sup_{\Db \in \Cfrak, \coeff \in \mathcal{K}}\|\Db \coeff\|_{2} < \infty$ \cite[Theorem 2]{MaurerPontil}, or of $\ell^{1}$ or $\ell^{0}$ balls \cite{vainsencher:2010}. Indeed, the first generic case dealt with in this paper involves penalty functions that only need to be non-negative, lower semi-continuous, coercive (and for convenience be minimum at zero). The second generic case also covers as particular cases the indicator function of $k$-sparse vectors and that of non-negative vectors, with restrictions on $\Cfrak$.

Beyond sparse coding, our results provide apparently new sample complexity estimates for $K$-means. Compared to state of the art results \cite{bartlett1998minimax} and \cite{biau2008performance} we pay an additional $\log \nsamp$ factor in our excess risk analysis, but improve the sample complexity estimate when the number of clusters $K$ exceeds the ambient dimension $\sdim$. A brief discussion on possible strategies of how to obtain results independent of $\sdim$ is provided in Section~\ref{sec:sample_alt}.

Similarly, our results provide new sample complexity estimates for NMF improving over state of the art results \cite{MaurerPontil} unless the ambient dimension $\sdim$ exceeds the order $\cdim^{2}$. Again, we refer to Section~\ref{sec:sample_alt} for a discussion of the high-dimensional setting.


Despite its successes, the main limitation of the proposed approach seems to be in the rather crude use of union bounds and covering numbers. One can envision  sharper sample complexity estimates using more refined concentration tools. For example, sample complexity results for PCA for signal distributions in the unit ball were studied in \cite{blanchard:2007}. Although our approach achieves results for more general distributions such as the class $\Pfrak_{A}$, the sample complexity results we obtain for this setting seem disappointingly pessimistic, both paying an additional $\log \nsamp$ factor and overestimating the sample complexity as $\sdim \cdim - \cdim(\cdim+1)/2$ rather than $\cdim$. We expect this may be due to the use by the authors in \cite{blanchard:2007} of more refined tools such as McDiarmid's bounded difference concentration inequality for the supremum of the deviation, and more importantly of a clever dimension-independent bound (i.e., independent of $\sdim$) on the expectation of the supremum of the deviation. 
We dedicated Section~\ref{sec:sample_alt} to a brief discussion of how we would envision to apply these tools to the more general context considered here, to obtain dimension-independent results.

The reader may have noticed that the rate of convergence of our estimates is in $1/\sqrt{\nsamp}$, which is typical when using techniques based on empirical processes. In certain settings it is possible with substantially more work to achieve fast rates in $1/\nsamp$, see, e.g., \cite{Sridharan:2008us} in a much simpler setting, or the discussions of fast rates in \cite{MaurerPontil,vainsencher:2010,biau2008performance,blanchard:2007}. It is not clear at this stage whether it is a realistic objective to achieve fast rates in the investigated general setting. In any case this is expected to require well chosen ``margin conditions'' to hold, see, e.g., \cite{Levrard:2013ho,Levrard:2014vq} for the case of $K$-means.

Improving the estimate of Lipschitz constants is another avenue for improvement, although its role in our sample complexity estimates is already only logarithmic. 
In fact, the generality of our approach comes at a price.
The bound on $L_{\Signal}(\Db)$ in Lemma~\ref{le:ConstantsWithNormPenalties_A1A3} (resp. Lemma~\ref{le:ConstantsWithKSparse}) is a worst case estimate given the bounds~\eqref{eq:residualBoundCrude_A1A3} and~\eqref{eq:coeffBound2_A1A3} (resp.~\eqref{eq:IntermediateEstimateCoeffKSparse}), and therefore rather crude. In a probabilistic setting, where the training samples $\signal_{i}$ are drawn i.i.d.\ according to some distribution $\PP$, one can envision much sharper bounds for $\tfrac{1}{\nsamp} \|(\Signal-\Db\Coeff)\Coeff^{\top}\|_\star$ using matrix concentration inequalities. A key difficulty will come from the control of $\Afrak$ which is dependent on $\Signal$ and $\Db$. Moreover, while we concentrated on the exploitation of Lemma~\ref{le:OneSidedLipschitz} and its corollary for the special metric $\|\cdot\|_{1 \to 2}$ on dictionaries, better choices may be possible, including metrics defined in terms of the penalty $g$ when applicable (potentially leading to alternate definitions of $\bar{g}$), or metrics dependent on the data distribution $\PP$.

The reader may object that the role of the Lipschitz constant $L$ in our final results (see, e.g., Theorem~\ref{th:MainTheorem}) is only logarithmic so the added value of such technicalities might be limited. 
Here, we notice that for the sake of simplicity we expressed Corollary~\ref{le:ConstantsWithNormPenalties} with a uniform Lipschitz constant independent of $\Db$. Clearly, some intermediate steps yield finer estimates $L_{\Signal}(\Db)$ that depend on the considered dictionary $\Db$, i.e., {\em local} Lipschitz constants. In certain scenarii these may provide more precise estimates that may turn out to be useful especially in the analysis of local properties of $F_{\Signal}(\Db)$. This is best illustrated with an example.

\begin{example}
When $g(\coeff) = \lambda \|\coeff\|_{1}^{2}$, we have $\bar{g}(t) = \sqrt{t/\lambda}$ hence $L_{\Signal}(\Db) = \tfrac{1}{\nsamp} \sum_{i=1}^{\nsamp} \sqrt{2f_{\signal_{i}}(\Db)} \cdot \sqrt{f_{\signal_{i}}(\Db)/\lambda} = F_{\Signal}(\Db) \sqrt{2/\lambda}$. As a result, Corollary~\ref{cor:LocalLipschitz} implies that we locally have $F_{\Signal}(\Db')/F_{\Signal}(\Db) \leq 1 + \sqrt{2/\lambda} \|\Db'-\Db\|_{1\to2}$. This implies that $\log F_{\Signal}(\Db)$ is uniformly Lipschitz with constant $L = \sqrt{2/\lambda}$. In other words, $F_{\Signal}(\Db)$ is more regular where it takes small values. \end{example}

This is likely to have an impact when performing a local stability analysis of sparse coding with the penalty $\lambda \|\coeff\|_{1}^{2}$ rather than $\lambda \|\coeff\|_{1}$, in the spirit of \cite{jenatton:hal-00737152}, see also \cite{Gribonval:2010aa,Geng:2011uq,schnass:2013}.

Finally, observing that this paper provides sample complexity estimates of matrix factorization with penalties $g(\coeff)$ much beyond indicator functions of (often compact) sets, it is natural to wonder if we could also replace the constraint $\Db \in \Cfrak$, i.e., the penalty $\chi_{\Cfrak}(\Db)$, by a more general penalty on $\Db$ promoting certain dictionaries. This would somehow lead to a model selection process where one could envision, e.g., adapting the effective number of nonzero columns of $\Db$, but would raise hard questions regarding the usability of covering numbers.

\section*{Acknowledgments}

R. Gribonval would like to thank Daniel Vainsencher and Karin Schnass for memorable discussions motivating this work and Gilles Blanchard for invaluable comments.


%

%

\appendices

\section{Proof of Lemma~\ref{lem:truncated_moments}}
To begin with, let us notice that by invoking twice the triangle inequality, we have 
\begin{align*}
	\left(\EE\left\{ \left| |y|^p  - \EE\left\{ |y|^p\right\}  \right|^q  \right\}\right)^{1/q}
	\leq {} &
	\left(\EE\left\{ |y|^{pq}\right\}\right)^{1/q} \\
	& + 
	\left(\EE\left\{ (\EE\left\{|y|^p\right\})^q \right\}\right)^{1/q}, 
\end{align*}
so that by using Jensen's inequality, we obtain
\begin{equation*}
	\EE\left\{ \left| |y|^p  - \EE\left[ |y|^p\right]  \right|^q  \right\}
	\leq
	2^q \EE\left[ |y|^{pq}\right],
\end{equation*}
thus proving~\eqref{eq:qMomentDeviation} provided that~\eqref{eq:qExpectation} holds. We now focus on these raw moments.
Let us fix some $u \geq 1$.
We introduce the event
\begin{equation*}
	\mathcal{K}\defin
	\left\{ \omega \,:\, |y(\omega)|/B \leq  u \right\},
\end{equation*}
and define $l_u$ as the largest integer such that $u \in [l_u,l_u+1)$.
We can then ``discretize'' the event $\mathcal{K}^c$ as
\begin{equation*}
	\mathcal{K}^c \subseteq \cup_{l=l_u}^\infty \mathcal{K}^c_l,\quad \text{with}\ 
	\mathcal{K}^c_l = 
	\left\{ \omega \,:\, 
	|y(\omega)|/B \in [l,l+1)\right\}.
\end{equation*}
We have
\begin{align*}
 \EE\{  |y|^{pq}\} &=
\EE\{ \indicator{\Kcal} |y|^{pq}\} + \EE\{ \indicator{\Kcal^c} |y|^{pq}\}\\
& \leq  \big(Bu\big)^{pq} + 
\textstyle\sum_{l=l_u}^\infty\EE\{ \indicator{\Kcal^c_l} |y|^{pq}\}\\
&\leq
B^{pq}
\cdot \Big[  u^{pq} + 
\textstyle\sum_{l=l_u}^\infty (l+1)^{pq} \cdot \EE\{ \indicator{\mathcal{K}^c_l}\} \Big]\\
&\leq
B^{pq}
\cdot \Big[u^{q} +  
\textstyle\sum_{l=l_u}^\infty (l+1)^{pq} \cdot \EE[ 
\indicator{\{ \omega;\
|y(\omega)| \geq B l \}} ] \Big]
\end{align*}
where in the last line we used $u^{p} \leq u$ since $u\geq 1$ and $p \leq 1$. Using the hypothesis~\eqref{eq:DefExponentialDecay}, we continue
\begin{equation*}
 \EE\{ |y|^{pq}\} \leq 
B^{pq} 
\cdot \Big[ u^{q}  +
\textstyle\sum_{l=l_u}^\infty (l+1)^{pq} \exp(-l) \Big].
\end{equation*}
Upper bounding the discrete sum by a continuous integral, 
we recognize here the incomplete Gamma function~\cite{Gautschi1998},
\begin{align*}
\textstyle\sum_{l=l_u}^\infty (l+1)^{pq} e^{-l}
&=
\textstyle\sum_{l=l_u}^\infty \int_{l}^{l+1}(l+1)^{pq} e^{-l} dt\\
&\leq
\textstyle\sum_{l=l_u}^\infty \int_{l}^{l+1}(t+1)^{pq} e^{-(t+1)+t+1-l} dt\\
&\leq
e^{2}
\textstyle\sum_{l=l_u}^\infty \int_{l}^{l+1}(t+1)^{pq} e^{-(t+1)} dt\\
&=
e^{2}
\textstyle\int_{l_{u}}^{\infty}(t+1)^{pq} e^{-(t+1)} dt\\
&=
e^{2}
\textstyle\int_{l_{u}+1}^{\infty}t^{pq} e^{-t} dt
\leq 
e^{2}
\int_{u}^{\infty}t^{q} e^{-t} dt\\
&=e^{2} \Gamma\left(q+1,u\right)
\end{align*}
where again we used $t^{pq}\leq t^{q}$ for $t \geq 1$. A standard formula~\cite[equation (1.3)]{Gautschi1998} leads to, for $u \geq 1$, 
\begin{equation*}
\Gamma(q+1,u)=q! \exp(-u) \textstyle\sum_{j=0}^q \frac{u^j}{j!} \leq e\, q! \exp(-u) u^q. 
\end{equation*}
Putting all the pieces together we reach the conclusion.

\section{Covering numbers}

To estimate covering numbers in Lemma~\ref{lem:cover} we used implicitly the following Lemma, which follows directly from the definition of covering numbers and will serve to extend this result to other constraint sets.
\begin{lemma}
\label{lem:cartesian_dic}
	Consider compact constraint sets $\Cfrak_1,\ldots, \Cfrak_k$ with respective covering number bounds $N_1,\ldots,N_k$ with respect to metrics $\rho_{1}(\cdot,\cdot),\ldots,\rho_{k}(\cdot,\cdot)$. 
	Then the covering number of the Cartesian product $\Cfrak \defin \Cfrak_1 \times \ldots \times \Cfrak_k$ (with respect to the $\max_{i} \rho_{i}(\Db_{i},\Db'_{i})$ metric) has the upper bound
	\begin{equation*}
		\Ncal(\Cfrak,\epsilon) \leq \prod_{i=1}^{k} N_i.
	\end{equation*}
\end{lemma}

In order to further extend these results to more elaborate constraint sets we resort to a result from \cite{Szarek:1997wd}.
\begin{lemma}
\label{lem:cov_number_map}
	Given a constraint set $\Cfrak$ with distance measure $\rho_1$, a normed space $\mathfrak{M}$ with distance measure $\rho$, and a mapping $\Phi\colon \mathfrak{M} \to \Cfrak$ that fulfill the conditions
	\begin{enumerate}
			\item There exists an $r \in \R{}, r>0$ such that $\Phi(B_r) \supset \Cfrak$ where $B_r$ is a ball around $0 \in \mathfrak{M}$ with radius $r$ with respect to the distance metric $\rho$ (Surjectivity).
			\item There exists an $L\in\R{}$ such that $\rho_1(\Phi(\Db_1),\Phi(\Db_2)) \leq L \cdot \rho(\Db_1, \Db_2)$ for $\Db_1,\Db_2 \in B_r$ (Lipschitz property).
	\end{enumerate}
	Then the covering number of $\Cfrak$ has the upper bound
	\begin{equation*}
		\Ncal(\Cfrak,\epsilon) \leq \left(3rL/\epsilon\right)^h
	\end{equation*}
	where the exponent $h$ is the dimension of the constraint set $\Cfrak$ in the sense of its manifold structure.
\end{lemma}

These lemmata imply covering number bounds for all the constraint sets introduced in the previous section.

\subsubsection{Sparse dictionaries}
The class of $s$-sparse dictionaries $\Cfrak_{\textrm{sparse}}(\sdim,\cdim,s,\mathbf{\Theta})$ with $\mathbf{\Theta} \in \Cfrak_{\textrm{orth}}(\sdim)$ is the Cartesian product of $\cdim$ copies of 
\begin{equation}
\label{eq:def_sparse_dic}
	\Cfrak_{\textrm{sparse}}(\sdim,1,s,\mathbf{\Theta}) \defin \{ \mathbf{\Theta} \mathbf{d}\,:\, \mathbf{d} \in \R{\cdim}, \|\mathbf{d}\|_{0} \leq s\}.
\end{equation}
By Lemma~\ref{lem:cartesian_dic} its covering number with the $\|\cdot\|_{1\to 2}$ metric is 
\begin{equation*}
	\Ncal(\Cfrak_{\mathrm{sparse}}(\sdim,\cdim,s,\mathbf{\Theta}),\epsilon) 
	\leq 
	\left( \Ncal(\Cfrak_{\mathrm{sparse}}(\sdim,1,s,\mathbf{\Theta}),\epsilon) \right)^\cdim.
\end{equation*}
Since the set $\Cfrak_{\mathrm{sparse}}(\sdim,1,s,\mathbf{\Theta})$ is isometric to $\Cfrak_{\mathrm{sparse}}(\sdim,1,s,\mathbf{I}_{\sdim})$, we have
$\Ncal(\Cfrak_{\mathrm{sparse}}(\sdim,1,s,\mathbf{\Theta}),\epsilon) = \Ncal(\Cfrak_{\mathrm{sparse}}(\sdim,1,s,\mathbf{I}_{\sdim}),\epsilon)$ these covering numbers are with respect to the Euclidean metric).
Now, observe that $\Cfrak_{\mathrm{sparse}}(\sdim,1,s,\mathbf{I}_{\sdim})$ is simply the set of normalized $s$-sparse vectors in $\R{\sdim}$, which is a union of 
$\tbinom{\sdim}{s}$ {$(s\!-\!1)$-spheres} from as many subspaces. Hence, we obtain
$\Ncal(\Cfrak_{\mathrm{sparse}}(\sdim,1,s,\mathbf{I}_{\sdim}),\epsilon) \leq \tbinom{\sdim}{s} \left(1+2/\epsilon\right)^{s}$, yielding
\begin{equation*}
	\Ncal(\Cfrak_{\mathrm{sparse}}(\sdim,\cdim,s,\mathbf{\Theta}),\epsilon) \leq \left( \tbinom{\sdim}{s} \left(1+2/\epsilon\right)^s \right)^\cdim.
\end{equation*}

\subsubsection{Orthogonal Dictionaries}
The orthogonal dictionaries as defined in Equation~\eqref{eq:def_Dorth} are the group of special orthogonal matrices. The special orthogonal group is a Lie group, and the corresponding Lie algebra is the vector space of skew symmetric matrices $\mathfrak{so}(\cdim)$, i.e., $\exp\colon \mathfrak{so}(\cdim) \to \Cfrak_\mathrm{orth}(\cdim)$ is a surjective map. Furthermore, it can be shown that it is sufficient to restrict $\mathfrak{so}(\cdim)$ to a ball around $\mathbf{0}$ with radius $\pi$ in order to obtain a surjective map. 
The exponential mapping is Lipschitz since it holds that 
\begin{equation}
\label{eq:Lipschitz_exp}
	\rho(\exp(\mathbf{S}_1),\exp(\mathbf{S}_2)) \leq e^a\|\mathbf{S}_1 - \mathbf{S}_2\|,
\end{equation}
with $\mathbf{S}_1, \mathbf{S}_2 \in B_\pi$ and $a = \max \left(\|\mathbf{S}_1\|,\|\mathbf{S}_2\| \right) \leq \pi$ and we obtain the Lipschitz constant $L=e^\pi$.
Finally, the dimension of $\Cfrak_\mathrm{orth}(\cdim)$ is $\cdim(\cdim-1)/2$ which yields the covering number bound
\begin{equation*}
	\Ncal(\Cfrak_\mathrm{orth}(\cdim),\epsilon) \leq \left( 3\pi e^\pi/\epsilon\right)^{\cdim(\cdim-1)/2}.
\end{equation*}

\subsubsection{Stiefel matrices}
The Stiefel manifold, defined in Equation~\eqref{eq:def_stiefel}, can also be defined as the quotient space $\SO(\sdim)/\SO(\sdim-\cdim)$, cf.~\cite{edelman:1998geometry}. In the previous section, we already mentioned that the Lie algebra of $\SO(\sdim)$ is given by $\mathfrak{so}(\sdim)$. Now, we define the function 
\begin{equation*}\begin{split}
	&\Phi\colon \mathfrak{so}(\sdim) \to \ST(\cdim,\sdim),\\
	&\Phi \defin q \circ \exp,
\end{split}\end{equation*}
where $q\colon \SO(\sdim) \to \SO(\sdim)/\SO(\sdim-\cdim)$ denotes the quotient mapping. 
The Lipschitz constant for the exponential mapping has been previously established in Equation~\eqref{eq:Lipschitz_exp}.
The function $q$ is a quotient map and therefore a contraction, and as the mapping $\Phi$ is a combination of two Lipschitz mappings it is Lipschitz itself. Thus, conditions \textsl{(1)} and \textsl{(2)} in Lemma~\ref{lem:cov_number_map} are fulfilled and taking into account that $\dim(\ST(\cdim,\sdim)) = \sdim \cdim - \cdim(\cdim+1)/2$ we obtain the covering number bound
\begin{equation*}
	\Ncal(\Cfrak_{\mathrm{St}}(\cdim,\sdim),\epsilon) \leq \left( 3\pi e^\pi/\epsilon \right)^{\sdim \cdim - \cdim(\cdim+1)/2}.
\end{equation*}

\subsubsection{Separable Dictionaries}
The mapping 
\begin{equation*}\begin{split}
	\Phi \colon &\Cfrak(\sdim_1,\cdim_1) \times \ldots \times \Cfrak(\sdim_z,\cdim_z)
	\to \Cfrak_{\mathrm{sep}},\\
	{}&(\Db_1,\Db_2,\ldots,\Db_2) \mapsto \Db_1 \otimes \Db_2 \otimes \ldots \otimes \Db_z
\end{split}\end{equation*}
is surjective and Lipschitz with a Lipschitz constant smaller than one. Thus, Lemmata~\ref{lem:cov_number_map}~\&~\ref{lem:cartesian_dic} yield
\begin{equation*}
	\Ncal(\Cfrak_{\mathrm{sep}},\epsilon) \leq \left( 3/\epsilon \right)^{\sum_i\sdim_i \cdim_i}.
\end{equation*}

\subsubsection{Tensor Product of Stiefel matrices}
Analogous to above we introduce the mapping 
\begin{equation*}\begin{split}
	\Phi \colon {}&\ST(\sdim_1,\cdim_1) \times \ldots \times \ST(\sdim_z,\cdim_z) 
	\to \Cfrak_{\otimes \mathrm{St}},\\
	{}&(\Db_1,\ldots,\Db_z) \mapsto \Db_1 \otimes \ldots \otimes \Db_z.
\end{split}\end{equation*}
As this is just a special case of the mapping in the previous constraint set, this mapping is surjective as well as a contraction. Hence, according to Lemmata~\ref{lem:cov_number_map} and\ref{lem:cartesian_dic} the covering number bound is given by
\begin{equation*}
	\Ncal(\Cfrak_{\otimes \mathrm{St}},\epsilon) \leq \left( 3\pi e^\pi/\epsilon \right)^{\sum_i \sdim_i \cdim_i - \cdim_i (\cdim_i +1)/2}.
\end{equation*}

\ifCLASSOPTIONcaptionsoff
  \newpage
\fi







\begin{thebibliography}{10}
\providecommand{\url}[1]{#1}
\csname url@samestyle\endcsname
\providecommand{\newblock}{\relax}
\providecommand{\bibinfo}[2]{#2}
\providecommand{\BIBentrySTDinterwordspacing}{\spaceskip=0pt\relax}
\providecommand{\BIBentryALTinterwordstretchfactor}{4}
\providecommand{\BIBentryALTinterwordspacing}{\spaceskip=\fontdimen2\font plus
\BIBentryALTinterwordstretchfactor\fontdimen3\font minus
  \fontdimen4\font\relax}
\providecommand{\BIBforeignlanguage}[2]{{%
\expandafter\ifx\csname l@#1\endcsname\relax
\typeout{** WARNING: IEEEtran.bst: No hyphenation pattern has been}%
\typeout{** loaded for the language `#1'. Using the pattern for}%
\typeout{** the default language instead.}%
\else
\language=\csname l@#1\endcsname
\fi
#2}}
\providecommand{\BIBdecl}{\relax}
\BIBdecl

\bibitem{Mallat:2008aa}
S.~Mallat, \emph{{A Wavelet Tour of Signal Processing}}, 3rd~ed.\hskip 1em plus
  0.5em minus 0.4em\relax Academic Press, 2008.

\bibitem{sp:elad:2010}
M.~Elad, M.~A.~T. Figueiredo, and Y.~M., ``On the role of sparse and redundant
  representations in image processing,'' \emph{Proceedings of the IEEE},
  vol.~98, no.~6, pp. 972--982, 2010.

\bibitem{Elad:2010wo}
M.~Elad, \emph{Sparse and Redundant Representations}, ser. From Theory to
  Applications in Signal and Image Processing.\hskip 1em plus 0.5em minus
  0.4em\relax Springer, 2010.

\bibitem{tucker:1966}
L.~R. Tucker, ``Some mathematical notes on three-mode factor analysis,''
  \emph{Psychometrika}, vol.~31, no.~3, pp. 279--311, 1966.

\bibitem{Olshausen:1997wz}
B.~A. Olshausen and D.~J. Field, ``{Sparse coding with an overcomplete basis
  set: A strategy employed by VI?}'' \emph{Vision Research}, vol.~37, no.~23,
  pp. 3311--3326, 1997.

\bibitem{dl:engan:1999}
K.~Engan, S.~Aase, and J.~Hakon~Husoy, ``Method of optimal directions for frame
  design,'' in \emph{IEEE International Conference on Acoustics, Speech, and
  Signal Processing}, 1999, pp. 2443--2446.

\bibitem{dl:delgado:2003}
K.~Kreutz-Delgado, J.~F. Murray, B.~D. Rao, K.~Engan, T.~W. Lee, and T.~J.
  Sejnowski, ``Dictionary learning algorithms for sparse representation,''
  \emph{Neural Computation}, vol.~15, no.~2, pp. 349--396, 2003.

\bibitem{dl:aharon:2006}
M.~Aharon, M.~Elad, and A.~Bruckstein, ``{K-SVD}: An algorithm for designing
  overcomplete dictionaries for sparse representation,'' \emph{IEEE
  Transactions on Signal Processing}, vol.~54, no.~11, pp. 4311--4322, 2006.

\bibitem{dl:mairal:2010}
J.~Mairal, F.~Bach, J.~Ponce, and G.~Sapiro, ``Online learning for matrix
  factorization and sparse coding,'' \emph{Journal of Machine Learning
  Research}, vol.~11, no.~1, pp. 19--60, 2010.

\bibitem{dl:tosic:2011}
I.~To\v{s}i\'c and P.~Frossard, ``Dictionary learning,'' \emph{IEEE Signal
  Processing Magazine}, vol.~28, no.~2, pp. 27--38, 2011.

\bibitem{Rubinstein:2010aa}
R.~Rubinstein, A.~M. Bruckstein, and M.~Elad, ``Dictionaries for sparse
  representation modeling,'' \emph{Proceedings of the IEEE}, vol.~98, no.~6,
  pp. 1045--1057, 2010.

\bibitem{cvpr:hawe:2013}
S.~Hawe, M.~Seibert, and M.~Kleinsteuber, ``Separable dictionary learning,'' in
  \emph{IEEE Conference on Computer Vision and Pattern Recognition}, 2013.

\bibitem{dl:rubinstein:2010}
R.~Rubinstein, M.~Zibulevsky, and M.~Elad, ``Double sparsity: Learning sparse
  dictionaries for sparse signal approximation,'' \emph{IEEE Transactions on
  Signal Processing}, vol.~58, no.~3, pp. 1553--1564, 2010.

\bibitem{seung:2001algorithms}
D.~Seung and L.~Lee, ``Algorithms for non-negative matrix factorization,''
  \emph{Advances in Neural Information Processing Systems}, vol.~13, pp.
  556--562, 2001.

\bibitem{Gersho:1992wy}
A.~Gersho and R.~M. Gray, \emph{Vector Quantization and Signal
  Compression}.\hskip 1em plus 0.5em minus 0.4em\relax Springer, 1992.

\bibitem{dAspremont:2004vv}
A.~d'Aspremont, L.~El~Ghaoui, M.~I. Jordan, and G.~R.~G. Lanckriet, ``{A direct
  formulation for sparse {PCA} using semidefinite programming},'' \emph{arXiv},
  Jun. 2004.

\bibitem{witten:2009penalized}
D.~M. Witten, R.~Tibshirani, and T.~Hastie, ``A penalized matrix decomposition,
  with applications to sparse principal components and canonical correlation
  analysis,'' \emph{Biostatistics}, vol.~10, no.~3, pp. 515--534, 2009.

\bibitem{Zhang:2012uu}
Y.~Zhang and L.~El~Ghaoui, ``Large-scale sparse principal component analysis
  with application to text data,'' \emph{arXiv}, Oct. 2012.

\bibitem{MaurerPontil}
A.~Maurer and M.~Pontil, ``K-dimensional coding schemes in hilbert spaces,''
  \emph{IEEE Transactions on Information Theory}, vol.~56, no.~11, pp.
  5839--5846, 2010.

\bibitem{vainsencher:2010}
D.~Vainsencher, S.~Mannor, and A.~M. Bruckstein, ``The sample complexity of
  dictionary learning,'' \emph{Journal of Machine Learning Research}, vol.~12,
  pp. 3259--3281, 2011.

\bibitem{jenatton:inria-00377732}
R.~Jenatton, J.-Y. Audibert, and F.~Bach, ``Structured variable selection with
  sparsity-inducing norms,'' \emph{Journal of Machine Learning Research},
  vol.~12, pp. 2777--2824, 2011.

\bibitem{chartrand:2007exact}
R.~Chartrand, ``Exact reconstruction of sparse signals via nonconvex
  minimization,'' \emph{IEEE Signal Processing Letters}, vol.~14, no.~10, pp.
  707--710, 2007.

\bibitem{mailhe:2008shift}
B.~Mailh{\'e}, S.~Lesage, R.~Gribonval, F.~Bimbot, P.~Vandergheynst
  \emph{et~al.}, ``Shift-invariant dictionary learning for sparse
  representations: extending {K-SVD},'' in \emph{16th EUropean SIgnal
  Processing COnference (EUSIPCO'08)}, 2008.

\bibitem{Jenatton:2011vv}
R.~Jenatton, J.~Mairal, G.~Obozinski, and F.~Bach, ``Proximal methods for
  hierarchical sparse coding,'' \emph{Journal of Machine Learning Research},
  vol.~12, pp. 2297--2334, 2011.

\bibitem{de2000multilinear}
L.~De~Lathauwer, B.~De~Moor, and J.~Vandewalle, ``A multilinear singular value
  decomposition,'' \emph{SIAM Journal on Matrix Analysis and Applications},
  vol.~21, no.~4, pp. 1253--1278, 2000.

\bibitem{jenatton:hal-00737152}
R.~Jenatton, R.~Gribonval, and F.~Bach, ``{Local stability and robustness of
  sparse dictionary learning in the presence of noise},'' CMAP, Tech. Rep.,
  2012.

\bibitem{Buldygin2000}
V.~V. Buldygin and I.~U.~V. Kozachenko, \emph{Metric characterization of random
  variables and random processes}.\hskip 1em plus 0.5em minus 0.4em\relax
  American Mathematical Society, 2000, vol. 188.

\bibitem{Spielman:2012ue}
D.~A. Spielman, H.~Wang, and J.~Wright, ``{Exact recovery of sparsely-used
  dictionaries},'' \emph{Journal of Machine Learning Research: Workshop and
  Conference Proceeedings; 25th Annual Conference on Learning Theory}, vol.~23,
  pp. 1--18, 2012.

\bibitem{Arora:2013vq}
S.~Arora, R.~Ge, and A.~Moitra, ``{New Algorithms for Learning Incoherent and
  Overcomplete Dictionaries},'' \emph{arXiv}, Aug. 2013.

\bibitem{Agarwal:2013tya}
A.~Agarwal, A.~Anandkumar, P.~Jain, P.~Netrapalli, and R.~Tandon, ``{Learning
  Sparsely Used Overcomplete Dictionaries via Alternating Minimization},''
  \emph{arXiv}, Oct. 2013.

\bibitem{Cucker2002}
F.~Cucker and S.~Smale, ``On the mathematical foundations of learning,''
  \emph{Bulletin of the American Mathematical Society}, vol.~39, pp. 1--49,
  2002.

\bibitem{CS:candes:2006a}
E.~J. Cand\`es, J.~Romberg, and T.~Tao, ``Robust uncertainty principles:
  {E}xact signal reconstruction from highly incomplete frequency information,''
  \emph{IEEE Transactions on Information Theory}, vol.~52, no.~2, pp. 489--509,
  2006.

\bibitem{Negahban:2012en}
S.~N. Negahban, P.~Ravikumar, M.~J. Wainwright, and B.~Yu, ``A unified
  framework for high-dimensional analysis of $m$-estimators with decomposable
  regularizers,'' \emph{Statistical Science}, vol.~27, no.~4, pp. 538--557,
  Nov. 2012.

\bibitem{Boyd2004}
S.~P. Boyd and L.~Vandenberghe, \emph{Convex Optimization}.\hskip 1em plus
  0.5em minus 0.4em\relax Cambridge University Press, 2004.

\bibitem{hage:robust2014}
C.~Hage and M.~Kleinsteuber, ``Robust {PCA} and subspace tracking from
  incomplete observations using $\ell_0$-surrogates,'' \emph{Computational
  Statistics}, vol.~29, no.~2, 2014.

\bibitem{Vershynin2010}
R.~Vershynin, ``Introduction to the non-asymptotic analysis of random
  matrices,'' Preprint arXiv:1011.3027, Tech. Rep., 2010.

\bibitem{Hsu2011}
D.~Hsu, S.~M. Kakade, and T.~Zhang, ``A tail inequality for quadratic forms of
  subgaussian random vectors,'' \emph{Electronic Communications in
  Probability}, vol.~17, no.~52, pp. 1--6, 2012.

\bibitem{Bennett:1962vd}
G.~Bennett, ``Probability inequalities for the sum of independent random
  variables,'' \emph{Journal of the American Statistical Association}, 1962.

\bibitem{ShaweTaylor:2005eu}
J.~Shawe-Taylor, C.~K.~I. Williams, N.~Cristianini, and J.~Kandola, ``On the
  eigenspectrum of the gram matrix and the generalization error of
  kernel-{PCA},'' \emph{IEEE Transactions on Information Theory}, vol.~51,
  no.~7, pp. 2510--2522, Jul. 2005.

\bibitem{blanchard:2007}
G.~Blanchard, O.~Bousquet, and L.~Zwald, ``Statistical properties of kernel
  principal component analysis,'' \emph{Machine Learning}, vol.~66, no. 2-3,
  pp. 259--294, 2007.

\bibitem{Pollard:1982km}
D.~Pollard, ``A central limit theorem for $k$-means clustering,'' \emph{The
  Annals of Probability}, vol.~10, no.~4, pp. 919--926, Nov. 1982.

\bibitem{Linder:2001jl}
T.~Linder, ``On the training distortion of vector quantizers,'' \emph{IEEE
  Transactions on Information Theory}, vol.~46, no.~4, pp. 1617--1623, 2000.

\bibitem{bartlett1998minimax}
P.~L. Bartlett, T.~Linder, and G.~Lugosi, ``The minimax distortion redundancy
  in empirical quantizer design,'' \emph{{IEEE} Transactions on Information
  Theory}, vol.~44, no.~5, pp. 1802--1813, 1998.

\bibitem{Antos:2005ik}
A.~Antos, ``{Improved minimax bounds on the test and training distortion of
  empirically designed vector quantizers},'' \emph{IEEE Transactions on
  Information Theory}, vol.~51, no.~11, pp. 4022--4032, 2005.

\bibitem{Linder:1994il}
T.~Linder, G.~Lugosi, and K.~Zeger, ``Rates of convergence in the source coding
  theorem, in empirical quantizer design, and in universal lossy source
  coding,'' \emph{IEEE Transactions on Information Theory}, vol.~40, no.~6, pp.
  1728--1740, 1994.

\bibitem{biau2008performance}
G.~Biau, L.~Devroye, and G.~Lugosi, ``On the performance of clustering in
  hilbert spaces,'' \emph{{IEEE} Transactions on Information Theory}, vol.~54,
  no.~2, pp. 781--790, 2008.

\bibitem{Sridharan:2008us}
K.~Sridharan, S.~Shalev-Shwartz, and N.~Srebro, ``{Fast Rates for Regularized
  Objectives},'' \emph{Advances in Neural Information Processing Systems},
  vol.~21, pp. 1545--1552, 2008.

\bibitem{Levrard:2013ho}
C.~Levrard, ``{Fast rates for empirical vector quantization},''
  \emph{Electronic Journal of Statistics}, vol.~7, pp. 1716--1746, 2013.

\bibitem{Levrard:2014vq}
------, ``{Non Asymptotic Bounds for Vector Quantization in Hilbert Spaces},''
  \emph{Annals of Statistics}, vol.~43, pp. 592--619, Apr. 2015.

\bibitem{Gribonval:2010aa}
R.~Gribonval and K.~Schnass, ``Dictionary identification - {S}parse
  matrix-factorisation via $\ell_1$ minimisation,'' \emph{IEEE Transactions on
  Information Theory}, vol.~56, no.~7, pp. 3523--3539, 2010.

\bibitem{Geng:2011uq}
Q.~Geng, H.~Wang, and J.~Wright, ``On the local correctness of $\ell^1$
  minimization for dictionary learning,'' \emph{arXiv}, Jan. 2011.

\bibitem{schnass:2013}
K.~Schnass, ``On the identifiability of overcomplete dictionaries via the
  minimisation principle underlying {K-SVD},'' \emph{CoRR}, vol. abs/1301.3375,
  2013.

\bibitem{Gautschi1998}
W.~Gautschi, ``The incomplete {G}amma functions since {T}ricomi,'' in \emph{In
  Tricomi's Ideas and Contemporary Applied Mathematics, Atti dei Convegni
  Lincei, n.147, Accademia Nazionale dei Lincei}, 1998.

\bibitem{Szarek:1997wd}
S.~J. Szarek, ``Metric entropy of homogeneous spaces and finsler geometry of
  classical lie groups,'' in \emph{Quantum Probability}, ser. Banach Center
  Publications, vol.~43.\hskip 1em plus 0.5em minus 0.4em\relax Polish Academy
  of Sciences, 1998, pp. 395--410.

\bibitem{edelman:1998geometry}
A.~Edelman, T.~A. Arias, and S.~T. Smith, ``The geometry of algorithms with
  orthogonality constraints,'' \emph{{SIAM} Journal on Matrix Analysis and
  Applications}, vol.~20, no.~2, pp. 303--353, 1998.

\end{thebibliography}

\begin{IEEEbiographynophoto}{R{\'e}mi Gribonval}(FM'14)  is a Senior Researcher with Inria (Rennes, France), and the scientific leader of the PANAMA research group on sparse audio processing. A former student at  {\'E}cole Normale Sup{\'e}rieure (Paris, France), he received the Ph. D. degree in applied mathematics from Universit{\'e} de Paris-IX Dauphine (Paris, France) in 1999, and his Habilitation {\`a} Diriger des Recherches in applied mathematics from Universit{\'e} de Rennes~I (Rennes, France) in 2007. His research focuses on mathematical signal processing, machine learning, approximation theory and statistics, with an emphasis on sparse approximation, audio source separation and compressed sensing. 
\end{IEEEbiographynophoto}

\begin{IEEEbiographynophoto}{Rodolphe Jenatton} received the PhD degree from the Ecole Normale Superieure, Cachan, France, in 2011 under the supervision of Francis Bach and Jean-Yves Audibert. He then joined the CMAP at Ecole Polytechnique, Palaiseau, France, as a postdoctoral researcher working with Alexandre d?Aspremont. From early 2013 until mid 2014, he worked for Criteo, Paris, France, where he was in charge of improving the statistical and optimization aspects of the ad prediction engine. He is now a machine learning scientist at Amazon, Berlin, Germany. His research interests revolve around machine learning, statistics, (convex) optimization, (structured) sparsity and unsupervised models based on latent factor representations.
\end{IEEEbiographynophoto}

\begin{IEEEbiographynophoto}{Francis Bach} graduated from the Ecole Polytechnique, Palaiseau, France, in 1997. He received
the Ph.D. degree in 2005 from the Computer Science Division at the University of California, Berkeley. He is the leading researcher of the Sierra project-team of INRIA in the Computer Science Department of the Ecole Normale Supérieure, Paris, France. His research interests include machine learning, statistics, optimization, graphical models, kernel methods, and statistical signal processing. He is currently the action editor of the Journal of Machine Learning Research and associate editor
of IEEE Transactions in Pattern Analysis and Machine Intelligence.
\end{IEEEbiographynophoto}

\begin{IEEEbiographynophoto}{Martin Kleinsteuber} received his PhD in Mathematics from the University of W\"urzburg, Germany, in 2006.
After post-doc positions at National ICT Australia Ltd, the Australian National University, Canberra, Australia, and the University of W\"urzburg, he has been appointed assistant professor for geometric optimization and machine learning at the Department of Electrical Engineering and Information Technology, TU M\"unchen, Germany, in 2009.   
He won the SIAM student paper prize in 2004 and the Robert-Sauer-Award of the Bavarian Academy of Science in 2008
for his works on Jacobi-type methods on Lie algebras. 
His research interests are in the areas of statistical signal processing, machine learning and computer vision, with an emphasis on optimization and learning methods for sparse and robust signal representations and blind signal separation.
\end{IEEEbiographynophoto}

\begin{IEEEbiographynophoto}{Matthias Seibert} received the Dipl.-math. degree from the Julius-Maximilians-Universit{\"a}t W{\"u}rzburg, W{\"u}rzburg, Germany, in 2012.
He is currently pursuing the doctoral degree at the Department of Electrical and Computer Engineering, Technische Universität M{\"u}nchen,
Munich, Germany.
His current research interests include sparse signal models, geometric optimization, and learning complexity.
\end{IEEEbiographynophoto}

%
%
\vfill  

\end{document}